\documentclass[10pt,journal,compsoc]{IEEEtran}

\usepackage[utf8]{inputenc} 
\usepackage[T1]{fontenc}    
\usepackage{hyperref}       
\usepackage{url}            
\usepackage{booktabs}       
\usepackage{amsfonts}       
\usepackage{nicefrac}       
\usepackage{microtype}      
\usepackage{xcolor}         
\usepackage{amsmath}
\usepackage{graphicx}
\usepackage{multirow}
\usepackage{makecell}
\usepackage{threeparttable}
\usepackage[caption=false, font=normalsize, labelfont=sf, textfont=sf]{subfig}

\ifCLASSOPTIONcompsoc
  \usepackage[nocompress]{cite}
\else
  \usepackage{cite}
\fi

\ifCLASSINFOpdf

\else

\fi

\begin{document}
%
\title{Hierarchical Estimation for Effective and Efficient \\ Sampling Graph Neural Network }
%
%
%
%

\author{Yang~Li,
        Bingbing~Xu,
        Qi~Cao,
        Yige~Yuan,
        and~Huawei~Shen
\IEEEcompsocitemizethanks{\IEEEcompsocthanksitem 
Yang Li is with the Data Intelligence System Research Center, Institute of Computing Technology (ICT), Chinese Academy of Sciences (CAS), Beijing, 100190, China. \protect\\
E-mail: liyang21s@ict.ac.cn

\IEEEcompsocthanksitem 
Bingbing Xu is with the Data Intelligence System Research Center, Institute of Computing Technology (ICT), Chinese Academy of Sciences (CAS), Beijing, 100190, China. \protect\\
E-mail: xubingbing@ict.ac.cn

\IEEEcompsocthanksitem 
Qi Cao, Yige Yuan and Huawei Shen are with the Data Intelligence System Research Center, Institute of Computing Technology (ICT), Chinese Academy of Sciences (CAS), Beijing, 100190, China. \protect\\
E-mail: \{caoqi, yuanyige20z, shenhuawei\}@ict.ac.cn

}}

%
%

\markboth{Journal of \LaTeX\ Class Files,~Vol.~14, No.~8, August~2015}%
{Shell \MakeLowercase{\textit{et al.}}: Bare Demo of IEEEtran.cls for Computer Society Journals}
%



\IEEEtitleabstractindextext{%
\begin{abstract}

Improving the scalability of GNNs is critical for large graphs. Existing methods leverage three sampling paradigms including node-wise, layer-wise and subgraph sampling, then design unbiased estimator for scalability. However, the high variance still severely hinders GNNs’ performance. On account that previous studies either lacks variance analysis or only focus on a particular sampling paradigm, we firstly propose an unified node sampling variance analysis framework and analyze the core challenge "circular dependency" for deriving the minimum variance sampler, i. e., sampling probability depends on node embeddings while node embeddings can not be calculated until sampling is finished. Existing studies either ignore the node embeddings or introduce external parameters, resulting in the lack of a both efficient and effective variance reduction methods. Therefore, we propose the \textbf{H}ierarchical \textbf{E}stimation based \textbf{S}ampling GNN (HE-SGNN) with first level estimating the node embeddings in sampling probability to break circular dependency, and second level employing sampling GNN operator to estimate the nodes’representations on the entire graph. Considering the technical difference, we propose different first level estimator, i.e., a time series simulation for layer-wise sampling and a feature based simulation for subgraph sampling. The experimental results on seven representative datasets demonstrate the effectiveness and efficiency of our method.









\end{abstract}

}

\maketitle

\IEEEdisplaynontitleabstractindextext

%
\IEEEpeerreviewmaketitle

\ifCLASSOPTIONcompsoc
\IEEEraisesectionheading{\section{Introduction}\label{sec:introduction}}
\else
\section{Introduction}

\fi

%
%
%
%
\IEEEPARstart{G}{raph} neural network (GNN) \cite{kipf2016semi}, \cite{hamilton2017inductive}, \cite{velickovic2017graph}, \cite{wu2019simplifying}, \cite{asif2021graph} has achieved great success in handling graph structure data. By aggregating the information of their neighbours and performing nonlinear feature transformation iteratively, the power of GNN's capturing features has been proved to be impressive \cite{xu2018powerful}. Currently, various variants of GNN such as graph convolution network (GCN) \cite{kipf2016semi}, graphSAGE \cite{hamilton2017inductive} and graph attention network (GAT) \cite{velickovic2017graph} are widely used in numerous  scenarios, e.g., traffic prediction \cite{cui2019traffic}, \cite{wang2020traffic}, \cite{jiang2022graph}, recommender systems \cite{fan2019graph}, \cite{wu2019session}, \cite{wu2020graph}. However, traditional multi-layer GNNs require the entire graph to be loaded into memory at once, and the neighbour size of each node expands exponentially as the number of layers increases. Such two challenges compel the traditional GNNs to lose the scalability confronted with large-scale graph data. Even if the method of mini-batch training \cite{kipf2016semi} copes with these problems to a certain extent, the exponentially expansion of neighbours' size still makes mini-batch based GNNs powerless to handle the huge computational and storage costs.

Therefore, numerous researchers aim to design GNN models suitable for large-scale graphs \cite{rong2019dropedge}, \cite{wu2019simplifying}, \cite{bojchevski2020scaling}, \cite{rossi2020sign},  \cite{zeng2021decoupling}. Among these studies, employing Bernoulli sampling to reduce the size of neighbours involved in the information aggregation for each node is one of the most prevalent methods \cite{hamilton2017inductive}, \cite{chen2017stochastic}, \cite{chen2018fastgcn}, \cite{zou2019layer}, \cite{huang2018adaptive}, \cite{zeng2019graphsaint}, \cite{chiang2019cluster}, \cite{liu2020bandit}, \cite{cong2020minimal},  \cite{zhang2021biased}. Presently, most of the previous studies can be divided into three paradigms: node-wise sampling \cite{hamilton2017inductive}, \cite{chen2017stochastic}, \cite{liu2020bandit}, \cite{cong2020minimal},  \cite{zhang2021biased}, layer-wise sampling  \cite{chen2018fastgcn}, \cite{zou2019layer}, \cite{huang2018adaptive}, and subgraph sampling \cite{zeng2019graphsaint}, \cite{chiang2019cluster}. Specifically, node-wise sampling limits the size of the neighbours by sampling a fixed number of nodes for each node but this paradigm still faces exponentially increasing complexity. By comparison, layer-wise sampling and subgraph sampling breakthroughly limit the exponentially expanding neighbour's size to linear growth with layer numbers. Layer-wise sampling fixes the total numbers of sampled nodes in each layer while subgraph sampling decomposes the original graph into small subgraphs and learns representation on each subgraph respectively to control the calculation cost. At the aspect of implementation, layer-wise sampling methods do sampling before each iteration while subgraph sampling only partition the original large-scale graph at the very beginning of the whole training process. This technical difference allows layer-wise sampling to take into account all the neighbours of each node as much as possible but reduces the training efficiency when the graph is giant, subgraph sampling stands in the opposite apparently.


\begin{figure*}[htbp]
\centering
\subfloat[]{\includegraphics[width=5in]{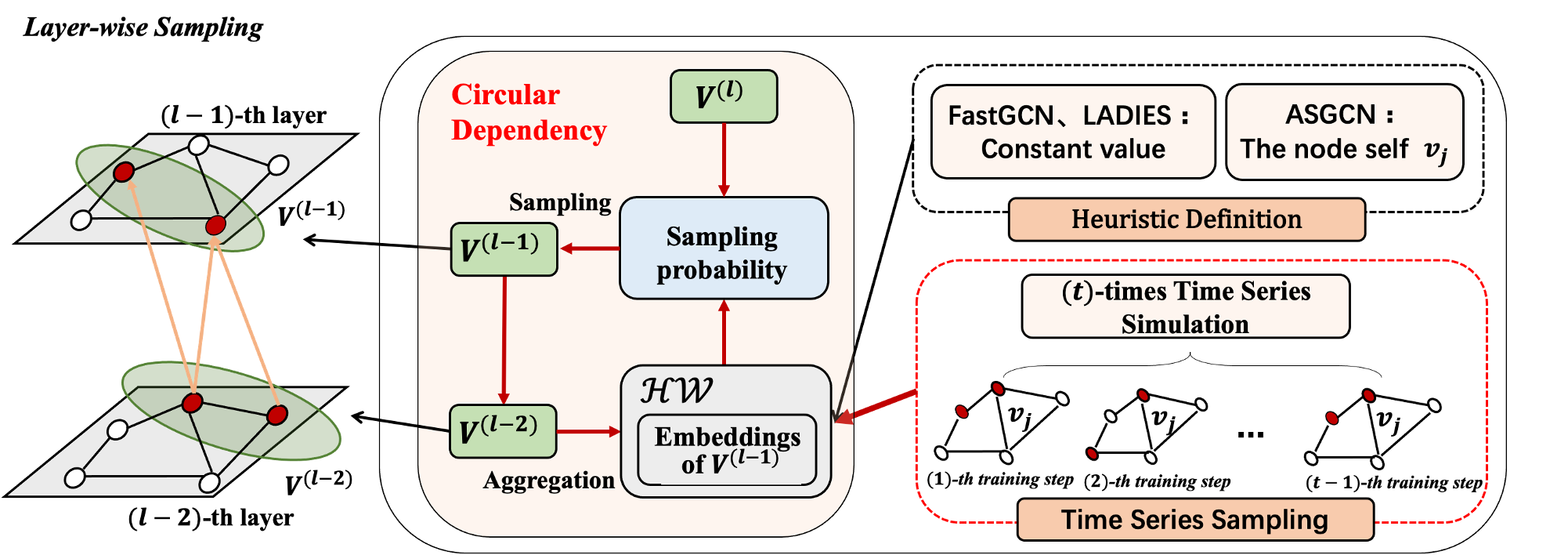}
\label{cd1}}

\subfloat[]{\includegraphics[width=3.3in]{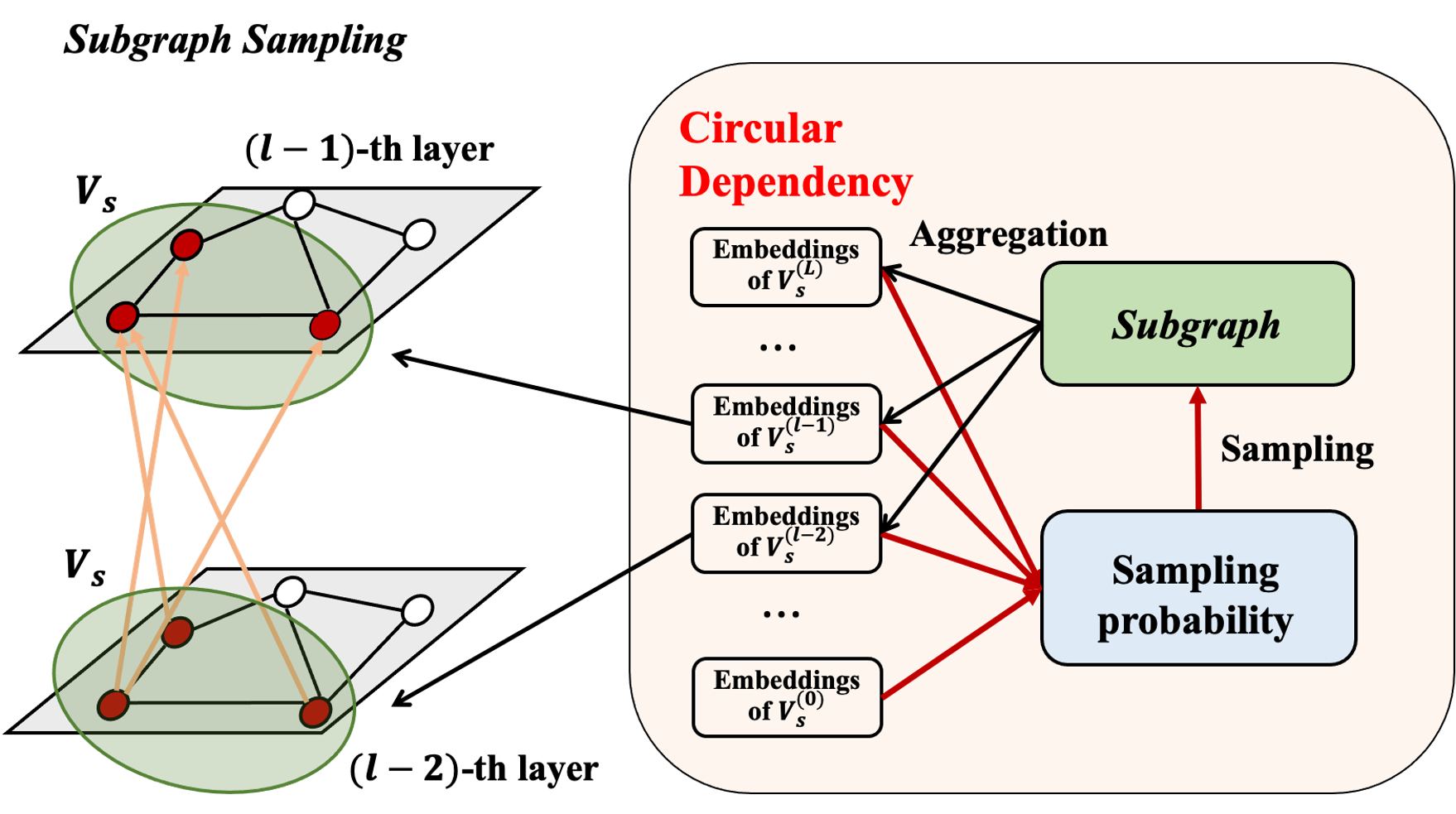}

\label{cd2}}
\subfloat[]{\includegraphics[width=3in]{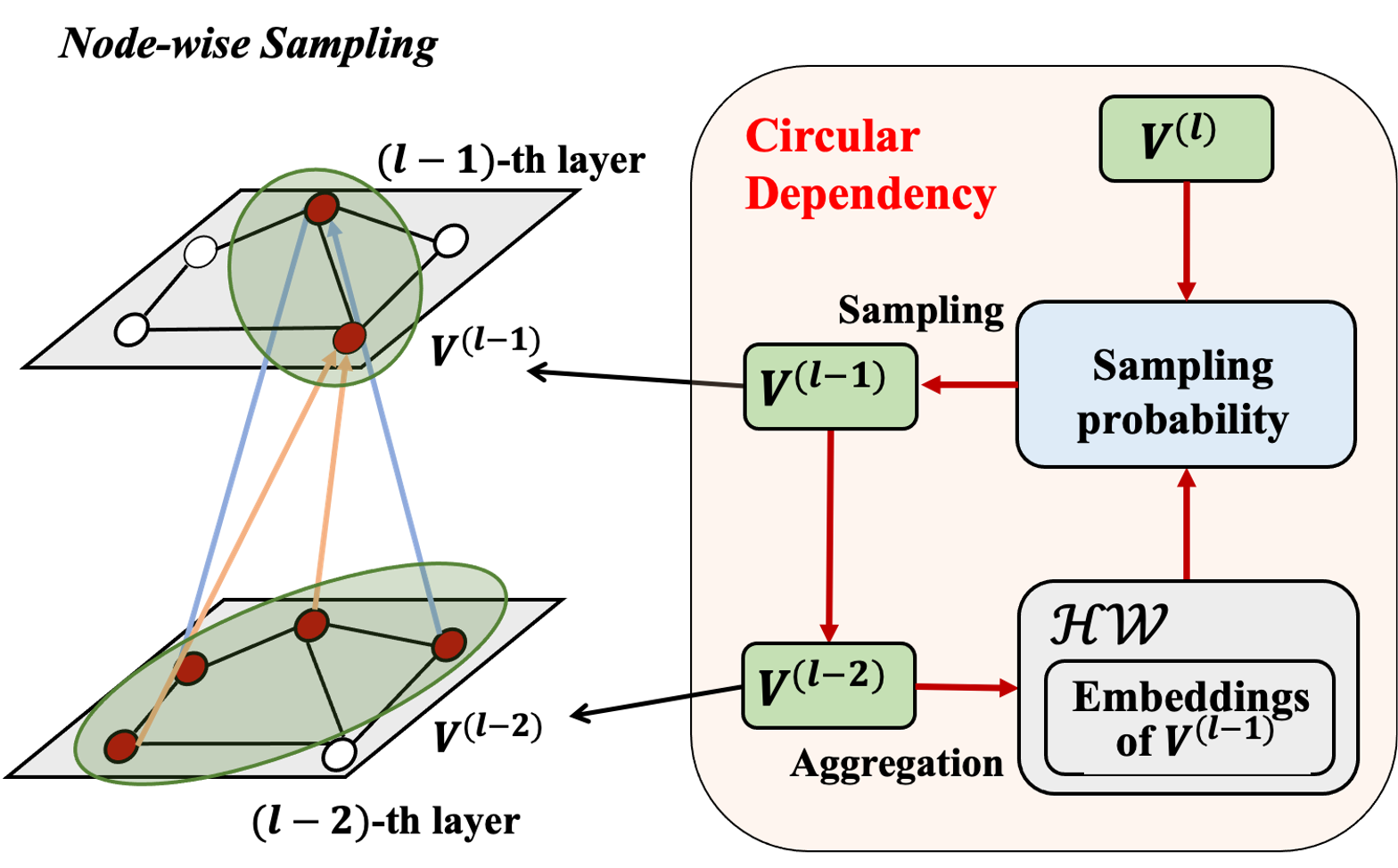}
\label{cd3}}
\caption{(a) Circular dependency in layer-wise sampling. (b) Circular dependency in subgraph sampling. (c) Circular dependency in node-wise sampling.}
\label{fig1}
\end{figure*}

The above paradigms can effectively reduce the cost of training while they will inevitably bring bias and variance to the traditional full-batch information aggregation operator. Significantly, excessive bias and variance will seriously lead to the nodes' representations learned by the model fluctuating on the computation graph sampled in different iteration, which hinders the effective training of graph neural networks. Several methods have successfully avoided bias to the greatest extent with the theoretically designed unbiased node samplers. However, it is difficult to reduce the variance of the node sampler on the basis of the unbiased estimation. Previous studies either ignore variance analysis such as ClusterGCN \cite{chiang2019cluster} and GraphSAINT \cite{zeng2019graphsaint} node sampler, or design a variance analysis framework limited to one of the three sampling paradigms, e. g., VRGCN focuses on the node-wise sampling, FastGCN \cite{chen2018fastgcn} and ASGCN \cite{huang2018adaptive} conduct variance analysis in layer-wise sampling. Taking into account the pros and cons of each sampling paradigm, a framework that can perform unified node sampling variance analysis for variance reduction on the three sampling paradigms is expected to be designed.

So in our study, we design an unified node sampling variance analysis for three above sampling paradigms to derive the minimum variance node sampler firstly. Then we analyze the critical challenge for obtaining minimum variance node sampler named circular dependency and point out the shortcomings of previous methods from the perspective of circular dependency. Next, considering the remaining exponent explosion in node-wise sampling, we propose the \textbf{H}ierarchical \textbf{E}stimation based \textbf{S}ampling GNN (HE-SGNN) for layer-wise sampling and subgraph sampling respectively to combat the circular dependency. Note that the hierarchical estimation node sampler strategy is applicable for node-sampling. Finally, We evaluate the performance of our proposed method on multiple datasets to verify its effectiveness and efficiency. In the remainder of this section, we will introduce the unified node sampling variance analysis framework and our hierarchical estimation in more detail.


Specifically, the unified node sampling variance analysis framework is obtained on the basis of unbiased estimator for vanilla GNN. We minimize the variance sum of the each node's embeddings in each layer obtained by unbiased estimator and find out that the expected closed-form solution of minimum variance node sampler cannot be obtained due to the circular dependency shown in Fig. \ref{fig1}. The sampling probability of minimum variance node sampler is determined by \textbf{$2$-norm of the multiplication of the node embeddings and parameters}. However, the node embeddings in $\mathcal{HW}$ depends on the nodes sampled in the sampling process, which is not feasible without determining the sampling probability, resulting in circular dependency.

Such dilemma also occurs in previous methods while is not well solved. For layer-wise sampling, previous methods utilize importance sampling \cite{glynn1989importance}, \cite{owen2013monte} and heuristically approximate the $\mathcal{HW}$ as shown in Fig. \ref{cd1}, e.g., leveraging constant value in FastGCN \cite{chen2018fastgcn} or the features of the node itself with external parameters in ASGCN \cite{huang2018adaptive}. However, these studies all ignore neighbour-aggregation information to  when calculating the sampling probability, resulting in variance not being effectively reduced. Furthermore, the external introduced parameters lead to a substantial increase in the sampling time cost which is hard to implement when layers goes deeper. For subgraph sampling, few studies have analyzed the attainment of the minimum variance node sampler. ClusterGCN \cite{chiang2019cluster} applies clustering algorithm which brings in bias and variance and cost much time for large-scale graph. GraphSAINT \cite{zeng2019graphsaint} also ignores the variance analysis and lacks considering the neighbour-aggregation information $\mathcal{HW}$ for their proposed node samplers. Overall, there still lacks a both efficient and effective variance reduction methods.

Under such situation, it is critical to solve such dilemma and approximate the minimum variance node sampler. We propose hierarchical estimation based sampling GNN with firstly estimating the node embeddings and calculate sampling probability based on this estimations and secondly employing GNN operator to aggregate the information of the sampled computation graph to estimate the nodes' representations on the entire graph for breaking circular dependency. In detail, we apply different skillful methods to design the hierarchical estimation in layer-wise sampling and subgraph sampling considering their technical difference. 

For layer-wise sampling, HE-SGNN establishes the sequence relationships among epochs and regards the historical information of the $\mathcal{HW}$ as multiple time series simulations demonstrated in Fig. \ref{cd1} to estimate the current $\mathcal{HW}$. To make each node can be sampled and then obtain a precise estimation, we employ optimistic initialization to increase the early exploration of this model. Note that this hierarchical estimation node sampler strategy is applicable for approximating the minimum variance sampler of node-wise sampling paradigm as well. For subgraph sampling, sampling is only done once before training, resulting in that no historical information is available for sampling to break circular dependency. So our proposed HE-SGNN firstly utilize the input node features to replace the neighbour-aggregation information of hidden layers for estimation, then aggregates information for estimating the node embeddings on the entire graph. However, there is no guarantee for the connection between the sampled nodes which causes inadequate neighbour-aggregation information for isolated nodes. Different form node sampler, edge sampler eliminates the the isolated nodes by conducting the partition by adding the two nodes connected with the sampled edge to the subgraph. Inspired by the edge sampler proposed in GraphSAINT, we innovatively induce an edge sampler from the node sampling probability which is acquired from our unified variance analysis framework.

In conclusion, the main contributions of our study are as follows:

\textbf{(1)} We unify the variance reduction analysis framework of three sampling paradigms and point out the critical challenge for obtaining the minimum variance sampler, i.e., circular dependency.

\textbf{(2)} We propose the hierarchical estimation based layer-wise sampling GNN, which leverages time series sampling to estimate the $\mathcal{HW}$ and calculates node sampling probability, then conducts sampling GNN. 


\textbf{(3)} We estimate the $\mathcal{HW}$ with input node features to break circular dependency for further estimating node embeddings in GNN, i.e., hierarchical estimation based subgraph sampling GNN. Furthermore, we apply node sampling probability induced edge sampler based GNN to learn node representations.

\textbf{(4)} We conduct experiments on seven authoritative datasets and demonstrate the superiority of our proposed model over the powerful baselines.

\section{Related Work}

Graph representation learning has received sustained attention in recent years. Different from Euclidean data, the complex structure between each node in one graph leads relevant algorithms difficult to learn effectively. Traditional works such as Deepwalk \cite{perozzi2014deepwalk} and Node2Vec \cite{grover2016node2vec} learn node embeddings with serialization modeling while LINE \cite{tang2015line} captures node features by defining structural similarities. With the rapid development of deep learning, Applying neural network models which have stronger expressive ability for graph representation learning has become a research hotspot. Inspired by convolutional neural network (CNN) \cite{kalchbrenner2014convolutional}, \cite{defferrard2016convolutional}, one of the representative models of deep learning, Kipf et al. \cite{kipf2016semi} develops graph convolution network which has successfully kicked off the employment of graph neural network (GNN) on graph. 

However, the dramatic increase in the graph scale has challenged the effective and efficient training of GNN severely. As a result, numerous studies have been carried out to reduce the neighbour's size involved in the information aggregation of each node. For example, Dropedge \cite{rong2019dropedge} randomly removes a certain number of edges to speed up the training, and GAS \cite{fey2021gnnautoscale} processes mini-batch-based GNN with historical embeddings to prune the computation graph. In the meantime, several studies reduce the size of computation graphs by using sampling methods and have designed promising scalable GNN models innovatively \cite{hamilton2017inductive}, \cite{chen2017stochastic}, \cite{chen2018fastgcn}, \cite{zou2019layer}, \cite{huang2018adaptive}, \cite{zeng2019graphsaint}, \cite{chiang2019cluster}, \cite{liu2020bandit}, \cite{cong2020minimal},  \cite{zhang2021biased}. These sampling based methods demonstrate efficient training process and maintain the effective expressive ability of GNN. On account of competitive performance of these sampling based methods, sampling to design scalable GNN has become one of the mainstream methods rapidly. To conclude, there consists of three sampling paradigms, i.e., node-wise sampling, layer-wise sampling, and subgraph sampling.

\textbf{Studies on Node-wise Sampling.}
Sampling based GraphSAGE \cite{hamilton2017inductive} controls the number of neighbors involved in the embedding aggregation to a small constant for each node. This sampling strategy is unbiased but the small number of samples may increase variance. VRGCN \cite{chen2017stochastic} and MVSGNN \cite{cong2020minimal} makes variance analysis of node-wise sampling by focusing on the difference between the representation of history and that of present which reduces the variance successfully.  However, this paradigm still faces the problem of exponentially increase in neighbor's size, thus losing its scalability when the layers are stacked deepeer. 

\textbf{Studies on Layer-wise Sampling.}
Layer-wise sampling is developed with the fixed-size node sets sampled at each layer. However, the circular dependency hinders achieving the minimum variance node sampler. In view of this dilemma, FastGCN~\cite{chen2018fastgcn} detaches the sampling probability from the dependency on the norm of hidden embeddings. Nevertheless, FastGCN cannot guarantee that sampled node sets between each layer are densely connected, which may result in large variances. Therefore, LADIES \cite{zou2019layer} modified the adjacency matrix for alleviating the sparseness problem caused by independent layer sampling. Additionally, ASGCN~\cite{huang2018adaptive} approximated the norm in probability distribution with modified graph attention \cite{vaswani2017attention}, \cite{velickovic2017graph} to further improve the aggregation results. But the calculation of attention costs too much time for sampling.

\textbf{Studies on Subgraph Sampling.}
Subgraph sampling constructs several subgraphs from original graph and learns the embeddings of nodes in each subgraph respectively. In subgraph sampling paradigm, the sampling probability is shared in all layers is and sampling is conducted only once before the training. For example, ClusterGCN \cite{chiang2019cluster} applies graph clustering algorithms to partition the original graph into several subgraphs. Though easy to implement, ClusterGCN inevitably introduces bias in embeddings aggregation and suffers huge costs on pre-processing the graph partition. Different from ClusterGCN, GraphSAINT \cite{zeng2019graphsaint} employs several well-developed samplers consisting of classical node sampler and edge sampler for devision to maginificantly speed up the preprocessing. But the studies in GraphSAINT doesn't consider the key elements, i. e. node embeddings. 

\section{Preliminaries}

In order to conduct variance analysis on the unbiased sampling estimator of vanilla GNN, the operator of vanilla GNN and its unbiased sampling estimator will be introduced in this section for further derivation.

\subsection{Graph Neural Networks}

Given an undirected graph $G=(V,E)$ where $V$ denotes the set of vertices in graph $G$ and $E$ denotes the set of edges in graph $G$, the traditional aggregation of GNN can be expressed as follows:

\begin{equation}
    H^{(l)}=\sigma(\hat{A}H^{(l-1)}W^{(l)}), \ \ l = 1,\cdots,L
    \label{(1)}
\end{equation}

\noindent where $\hat{A} \in \mathbf{R}^{N \times N}$ denotes the degree normalized adjacency matrix $A$ of $G$; $N$ denotes the number of vertices; $\sigma$ denotes the nonlinear activation function; $W^{(l)} \in \mathbf{R}^{d^{(l-1)} \times d^{(l)}}$ denotes the parameter matrix in $(l)$-th layer; $H^{(l)}$ denotes the embeddings matrix of $(l)$-th layer and $H^{(0)}$ is the features matrix $X\in \mathbf{R}^{N \times d^{(0)}}$. Then the hidden representation $h^{(l)}(v_j)$ of node $v_i \in V$ in $(l)$-th layer can be demonstrated by

\begin{equation}
    h^{(l)}(v_i)=\sigma(\sum_{j \in N(v_{i})}\hat{A}_{ij}h^{(l-1)}(v_j)W^{(l)})
    \label{(2)}
\end{equation}

\subsection{Unbiased Sampling Estimator for Vanilla GNN}

The vanilla GNN needs all the neighbours to conduct message passing. Let $F^{(l)}(v_i)=\sum^{N}_{j=1}\hat{A}_{ij}h^{(l-1)}(v_j)W^{(l)}$, $F^{(l)}(v_i)$ can be rewritten into expectation for introducing sampling probability on the whole graph as

\begin{equation}
    F^{(l)}(v_i)=\mathbb{E}_{q_{j}^{(l-1)}}[\frac{\hat{A}_{ij}}{q_{j}^{(l-1)}}h^{(l-1)}(v_j)W^{(l)}]
    \label{(3)}
\end{equation}

where $q_{j}^{(l-1)}$ denotes the node sampling probability distribution in $(l-1)$-th layer. Then by materializing the sampling probability, we can apply corresponding sampling process to reduce the scale of neighbours involved in information aggregation. Furthermore, in order to maintain the operator of vanilla GNN, an unbiased estimator of $F^{(l)}(v_i)$ is required to be designed.








\textbf{Theorem 1.} (Proof in Appendix \ref{a1}) \textit{$\hat{F}^{(l)}(v_i)$ is an unbiased estimator of $F^{(l)}(v_i)$ when $\hat{F}^{(l)}(v_i)$ is given by}

\begin{equation}
    \hat{F}^{(l)}(v_i)=\frac{1}{n_{i}^{(l-1)}}\sum_{j \in V^{(l-1)}_{i}}\frac{\hat{A}_{ij}}{q_{j}^{(l-1)}}h^{(l-1)}(v_j)W^{(l)}
    \label{(4)}
\end{equation}

\noindent \textit{where $n_{i}^{(l-1)}$ is the sampled nodes' number for estimating the ground truth $F^{(l)}(v_i)$ and $V^{(l-1)}_{i}$ denotes the sampled nodes set for aggregating node $v_i$.}

Then the aggregation of sampling GNN with unbiased estimator is integrated as follows:

\begin{equation}
    h^{(l)}(v_i)=\sigma(\frac{1}{n_{i}^{(l-1)}}\sum_{j=1}^{n_{j}^{(l-1)}}\frac{\hat{A}_{ij}}{q_{j}^{(l-1)}}h^{(l-1)}(v_j)W^{(l)})
    \label{(5)}
\end{equation}

\section{Proposed Method: HE-SGNN}

In this section, we will demonstrate the unified variance analysis framework and point out the critical challenge for obtaining the closed-form minimum variance node sampler. Then we introduce the proposed \textbf{H}ierarchical \textbf{E}stimation based \textbf{S}ampling GNN (HE-SGNN) respectively for layer-wise sampling and subgraph sampling in detail.

\subsection{Unified Node Sampling Variance Analysis Framework}

Given one GNN structure with $L$ layers, the minimum variance node sampler minimizes the variance sum of the unbiased estimator $\hat{F}^{(l)}(v_i)$ for each node $v_{i}$ in all layers. 
The optimization objective is expressed as 

\begin{equation}
    \underset{q_{j}^{(l-1)}}{\mathrm{argmin}}\sum_{l=1}^{L}\sum_{i=1}^{N}{Var}_{q_{j}^{(l-1)}}(\hat{F}^{(l)}(v_i))
    \label{(6)}
\end{equation}

\textbf{Theorem 2.} (Proof in Appendix \ref{a2}) \textit{The variance of $\hat{F}^{(l)}(v_i)$ for node $v_{i}$ in $l$-th layer is given by}

\begin{equation}
    \begin{split}    
    &{Var}_{q_{j}^{(l-1)}}(\hat{F}^{(l)}(v_i)) \\
    &\ \ \ \ = \mathbb{E}_{q_{j}^{(l-1)}}[\frac{(\hat{A}_{ij}h^{(l-1)}(v_j)W^{(l)}-F^{(l)}(v_i)q_{j}^{(l-1)})^2}{n_{i}^{(l-1)}(q_{j}^{(l-1)})^2}]
    \end{split}
    \label{(7)}
\end{equation}

The intuitive solution is to derive the node sampling probability $q_{j}^{(l-1)}$ from Eq. (\ref{(7)}) to obtain the minimum variance node sampler. However, the node sampling probability must strictly satisfy the constraint that $\sum_{j=1}^{N}q_{j}^{(l-1)}=1$. So we introduce Lagrange multiplier \cite{rockafellar1993lagrange} to rewrite the optimization objective in Eq. (\ref{(6)}) into  

\begin{equation}
\small
    \underset{q_{j}^{(l-1)}}{\mathrm{argmin}}\sum_{l=1}^{L}[\sum_{i=1}^{N}{Var}_{q_{j}^{(l-1)}} (\hat{F}^{(l)}(v_i)) + \lambda_{l-1}(1-\sum_{j=1}^{N}q_{j}^{(l-1)})]
    \label{(8)}
\end{equation}

\noindent where $\lambda_{l-1}(l = 1,\cdots,L)$ denotes Lagrange multiplier. Eq. ({\ref{(8)}) unifies the variance analysis of the three sampling paradigms. By assigning different $q_{j}^{(l-1)}$, Eq. ({\ref{(8)}) follow different sampling paradigm:

\begin{equation}
\small
    \begin{cases}
        \text{node-wise},\ q_{j}^{(l-1)} = q^{(l-1)}(v_j |v_i) \\
        \text{layer-wise},\ q_{j}^{(l-1)} = q^{(l-1)}(v_j |{V}^{(l)}) \\ \text{subgraph},\ q_{j}^{(l-1)} = q(v_j)
    \end{cases}
    \label{(9)}
\end{equation}

\noindent where ${V}^{(l)}$ denotes the already sampled nodes set at $(l)$-th layer in laywe-wise sampling. The sampling distributions of each node in ${V}^{(l)}$ in each layer is different for node-wise sampling while the node in one layer shares equivalent sampling distribution in layer-wise sampling. Additionally, there is only one probability distribution to partition the subgraph. Specially, subgraph node sampling is usually regarded as an independent sampling for each node.

\begin{figure*}[t]
\centering
\includegraphics[width=6.in]{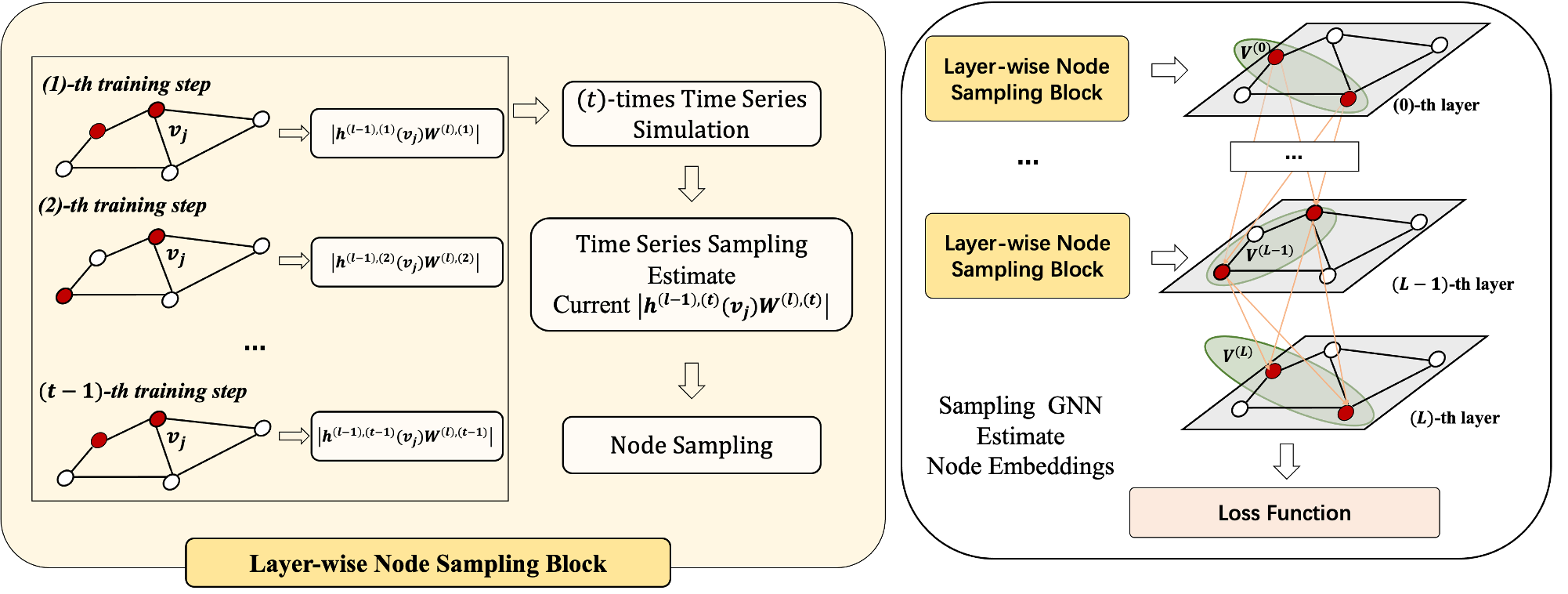}
\caption{Overall architecture of HE-SGNN with layer-wise node sampler.}
\label{layer}
\end{figure*}

\subsection{Scenario in Layer-wise Sampling}

\subsubsection{Circular Dependency in Layer-wise Sampling} 

Let $q_{j}^{(l-1)} = q^{(l-1)}(v_j |{V}^{(l)})$, we can derive the closed-form solution of minimum variance node sampler. Now we have the optimization objective function as: 

\begin{align}
        &L(\lambda_{1}, \lambda_{2},\cdots,\lambda_{L},q^{(l-1)}(v_j |{V}^{(l)}))\nonumber\\
        =& \sum_{l=1}^{L}[\sum_{i=1}^{N}{Var}_{q^{(l-1)}(v_j |{V}^{(l)})} (\hat{F}^{(l)}(v_i))\nonumber\\ 
        +& \lambda_{l-1}(1-\sum_{j=1}^{N}q^{(l-1)}(v_j |{V}^{(l)}))]
    \label{(10)}
\end{align}

Then we can find the extreme value of $L(\lambda_{1}, \lambda_{2},\cdots,\lambda_{L},q^{(l-1)}(v_j |{V}^{(l)}))$ by letting

\begin{equation}
    \begin{split}    
        &\frac{\partial L(\lambda_{1}, \lambda_{2},\cdots,\lambda_{L},q^{(l-1)}(v_j |{V}^{(l)}))}{\partial q^{(l-1)}(v_j|\hat{V}^{(l)})} = 0\\
        &\ \ \ s.\ t.\ \frac{\partial L(\lambda_{1}, \lambda_{2},\cdots,\lambda_{L},q^{(l-1)}(v_j |{V}^{(l)}))}{\partial \lambda_{l}} = 0
            \end{split}
    \label{(11)}
\end{equation}

With Eq. (\ref{(11)}), we can obtain the closed-form solution of layer-wise minimum variance node sampler of $(l)$-th layer. 

\textbf{Theorem 3.} (Proof in Appendix \ref{a3}) \textit{The probability distribution of layer-wise minimum variance node sampler is given by}

\begin{equation}
\small
    \begin{split}
        {q^{*}}^{(l-1)}(v_j |{V}^{(l)}) = \frac{(\sum_{i \in V^{(l)}}\hat{A}^{2}_{ij})^{\frac{1}{2}}|h^{(l-1)}(v_j)W^{(l)}|}{\sum_{k=1}^{N}(\sum_{i \in V^{(l)}}\hat{A}^{2}_{ik})^{\frac{1}{2}}|h^{(l-1)}(v_k)W^{(l)}|}
    \end{split}
    \label{(12)}
\end{equation}

The above closed-form of node sampler with minimum variance is determined by the node embeddings, parameter matrix and the adjacency matrix. Therefore, it cannot be solved due to the circular dependency (illustrated in Fig. \ref{cd1}). Specifically, the sampling probability of nodes in $(l-1)$-th layer $q^{(l-1)}$ depends on both $|h^{(l-1)}(v_j)W^{(l)}|$ and node set of $(l)$-th layer ${V}^{(l)}$. The node embeddings $h^{(l-1)}(v_j)$ in $|h^{(l-1)}(v_j)W^{(l)}|$ are aggregated from $(l-2)$-th layer. However, only after sampling the $(l-1)$-th layer node set, the $(l-2)$-th layer nodes can be sampled and obtained, resulting in the norm $|h^{(l-1)}(v_j)W^{(l)}|$ cannot be calculated. In other words, the top-down sampling process and the bottom-up aggregation process cause circular dependency.

Similarly, we can derive the minimum node-wise node sampler by replacing the $q^{(l-1)}(v_j|\hat{V}^{(l)})$ with $q^{(l-1)}(v_j |v_i)$ as ${q^{*}}^{(l-1)}(v_j |v_i)=\frac{A_{ij}|h^{(l-1)}(v_j)W^{(l)}|}{\sum_{k=1}^{N}A_{ik}|h^{(l-1)}(v_k)W^{(l)}|}$ and circular dependency also makes this closed-form solution hard to calculate.

\subsubsection{Motivation} 

In view of the existed circular dependency, various layer-wise sampling methods attempt to break this dilemma. Taking three representative methods as examples, FastGCN ignores $\mathcal{HW}$, i.e. \textbf{the multiplication result of the node embeddings and parameters}, and make the sampling probability independent of the nodes set of upper layers while LADIES replace this norm with constant, i.e., only considering $\hat{A}_{ij}$ to calculate sampling probability. In addition, ASGCN design the self-dependent function $|W_{g}h^{(0)}(v_j)|$ where $W_g$ denotes the external learned parameters. However, they all heuristically do not consider the aggregation information in $h^{(l-1)}(v_j)$. And the external parameters in ASGCN cause high time cost in the sampling process. Therefore, an efficient sampling method with a more precise approximation for variance minimum node sampler deserves to be designed.

\subsubsection{HE-SGNN with Layer-wise Node Sampler} 

In general, the proposed HE-SGNN with Layer-wise Node Sampler illustrated in Fig. \ref{layer} estimates the $\mathcal{HW}$ with well designed modules and then estimate the node embeddings in vanilla GNN with sampling GNN.

\textbf{Estimating the $\mathcal{HW}$.} Briefly, the key of circular dependency is $\mathcal{HW}$ which is unknown before top-down sampling process is completed. To solve this, we propose to use time series sampling to estimate $\mathcal{HW}$. Specifically, we find that node $v_j$ may be sampled in $(l-1)$-th layer before the current training epoch, i.e., the historical information of $\mathcal{HW}$ can be acquired. When the training is sufficient, the historical information of $\mathcal{HW}$ will be close to the current unknown representation. As a result, current unknown $\mathcal{HW}$ can be estimated with the historical representations.


So we innovatively treat each historical training as a sequence simulation for estimating current unknown $\mathcal{HW}$ shown in Fig. \ref{layer}. By establishing the sequential relationships among epochs and utilizing the sampling results from time series sampling, we estimate the current unknown $\mathcal{HW}$, thus successfully breaking the circular dependency without any additional parameters. Therefore, ${q^{(l-1)}(v_j|\hat{V}^{(l)})}^{*}$ can be approximated as

\begin{equation}
    {\hat{q}^{(l-1)}(v_j|\hat{V}^{(l)})}=\frac{(\sum_{i \in V^{(l)}}\hat{A}^{2}_{ij]})^{\frac{1}{2}}\mathbb{E}^{(T)}(\mathcal{HW}_{v_j})}{\sum_{k=1}^{N}(\sum_{i \in V^{(l)}}\hat{A}^{2}_{ik})^{\frac{1}{2}}\mathbb{E}^{(T)}(\mathcal{HW}_{v_k})}
    \label{(13)}
\end{equation}

where $T$ denotes the current iteration, $\mathbb{E}^{(T)}(\mathcal{HW}_{v_{j}})$ denotes the estimated expectation of $|h^{(l-1)}(v_j)W^{(l)}|$ at $T$-th iteration, which is calculated as follows: 

\begin{equation}
\footnotesize{
    \mathbb{E}^{(T)}(\mathcal{HW}_{v_{j}}) = \frac{\sum_{t=0}^{T-1}|h^{(l-1),(t)}(v_j)W^{(l),(t)}|\mathbb{I}(v_j,l-1,t)}{{n(v_j)}}}
    \label{(14)}
\end{equation}

where $n(v_j)$ denotes the times that $v_j$ is sampled in $(l-1)$-th layer in history and $\mathbb{I}(v_j,l-1,t)$ denotes indicator function. $\mathbb{I}(v_j,l-1,t) = 1$ when node $v_j$ is sampled in $(l-1)$-th layer at $t$-th iteration. Specially, $\mathbb{E}^{(1)}(\mathcal{HW}_{v_{j}})$ is the initialization term. Meanwhile, to prevent storing $\mathcal{HW}$ of all nodes at each epoch, the Eq. (\ref{(14)}) can be rewritten as Eq. (\ref{(15)}) via iterating:

\begin{equation}
\small
\begin{split}   
    &\mathbb{E}^{(T+1)}(\mathcal{HW}_{v_{j}}) \\
    &=\begin{cases}
    \mathbb{E}^{(T)}(\mathcal{HW}_{v_{j}}),\ \text{if}\  \mathbb{I}(v_j,l-1,t) = 0 \\
    \frac{ N^{(T)}(v_j) \mathbb{E}^{(T)}(\mathcal{HW}_{v_{j}}) + |h^{(l-1),(T)}(v_j)W^{(l),(T)}|}{ N^{(T)}(v_j) + 1},\  \text{else}\\
    \end{cases}
\end{split}
    \label{(15)}
\end{equation}

where $N^{(T+1)}(v_j) = N^{(T)}(v_j) + 1$ if $\mathbb{I}(v_j,l-1,t) = 1$ else $N^{(T+1)}(v_j) = N^{(T)}(v_j)$ and the initial term $\ N^{(1)}(v_j)=1$. Based on Eq. (\ref{(15)}), we can update the historical information via only one sum operation with almost no additional time complexity. On account that we only store the norm $\mathcal{HW}$ instead of the multiplication of the node embeddings and parameters, additional space complexity is $O(LN)$ ($L$ denotes the numbers of layers), which does not grow significantly.

\textbf{Estimating the Node Embeddings for Vanilla GNN.} Then we can estimate the node embeddings in vanilla GNN with proposed hierarchical estimation based layer-wise node sampling GNN demonstrated in Fig. \ref{layer} as 

\begin{equation}
\footnotesize
    h^{(l)}(v_i)=\sigma(\frac{1}{n_{i}^{(l-1)}}\sum_{j \in V^{(l-1)}}\frac{\hat{A}_{ij}}{\hat{q}^{(l-1)}(v_j|V^{(l)})}h^{(l-1)}(v_j)W^{(l)})
    \label{(16)}
\end{equation}

\noindent where $\hat{q}^{(l-1)}(v_j|V^{(l)})$ denotes the approximate sampling probability in layer-wise node sampling. Based on this two steps, we propose hierarchical estimation based layer-wise node sampling GNN illustrated in Fig. \ref{layer}, which including two parts: first using time series sampling to estimate $\mathcal{HW}$ and calculate the sampling probability, then conducting node sampling based on the probability to sample nodes and learn the representations for estimating the node embeddings in vanilla GNN.

\textbf{Optimistic Initialization.} We apply Optimistic Initialization for $\mathbb{E}^{(1)}(\mathcal{HW}_{v_{j}})$ of node $v_j$ to make the above Eq. (\ref{(15)}) to approximate $\mathcal{HW}$ more precisely. We expect this sampling GNN can accurately approximate the $\mathcal{HW}$ for all nodes. However, random initialization of the initial term $E^{(1)}(\mathcal{HW}_{v_j})$ will cause nodes with small initialized values fail to be sampled. Specifically, once a node is initialized to a small value, the calculated sampling probability (Eq. (\ref{(12)})) is small, making it difficult for the node to be sampled. So we apply optimistic initialization\cite{tijsma2016comparing}, i.e., initializing $E^{(1)}(\mathcal{HW}_{v_j})$ with larger values, to encourage the early exploration of this proposed model. The large initial value ensures the unsampled nodes with larger probability to be sampled, thus prompting them to participate in the embedding aggregation to approximate their actual $\mathcal{HW}$s.

\begin{table}
\renewcommand\arraystretch{1.75}
\scriptsize
\centering
\caption{Batch Training Time Complexity Comparison in Layer-wise Sampling GCNs}
\begin{tabular}{cp{2cm}cp{2cm}cp{1.8cm}}
\toprule
\multicolumn{1}{c}{Paradigm} &\multicolumn{1}{c}{Method} & \multicolumn{1}{c}{Time Complexity}\\
\midrule
\multicolumn{1}{c}{Mini Batched} &\multicolumn{1}{c}{GCN} & \multicolumn{1}{c}{$O(sD^Ld^2)$}\\
\midrule
\multirow{4}{*}{\makecell[b]{Layer-wise\\Sampling}} & \makecell[c]{FastGCN} & \makecell[c]{$O(L{||A_{s}||}_{0}d + Lsd^2)$}\\
 & \makecell[c]{LADIES} & \makecell[c]{$O(L{||A_{s}||}_{0}d + Lsd^2)$}\\
  & \makecell[c]{ASGCN} & \makecell[c]{$O(L{||A_{s}||}_{0}d + Lsd^2 + Ls^2d)$}\\
   & \makecell[c]{\textbf{HE-SGNN(L)}} & \makecell[c]{$O(L{||A_{s}||}_{0}d + Lsd^2)$}\\
\bottomrule
\end{tabular}
\label{t5}
\end{table}

\subsubsection{Time Copmplexity Analysis} 

Meanwhile, the efficiency of our HE-SGNN is also competitive. We analyze the batch training time complexity of prevalent layer-wise sampling methods in Table \ref{t5}. $s$ denotes the batch size; $A_{s}$ denotes the computation graph; $D$ denotes the average degree of the graph; $d$ denotes the dimension of features and $L$ denotes the total number of layers. Our layer-wise node sampler costs $O(Lsd)$ to calculate the $\mathcal{HW}$ which can be discarded as a small term. So the time complexity of our method is the same as the most effective models such as FastGCN and LADIES while ASGCN cost much time to calulate the attention which can not be ignored.

\subsection{Scenario in Subgraph Sampling}

\subsubsection{Circular Dependency in Subgraph Sampling} 

For subgraph sampling, the closed-form of minimum variance node sampler turns out to be more difficult to derive with $\ q_{j}^{(l-1)} = q(v_j)$, the optimization objective is demonstrated as:

\begin{align}
        &L(\lambda,q(v_j))\nonumber\\
        =&\sum_{l=1}^{L}\sum_{i=1}^{N}{Var}_{q(v_j)} (\tilde{F}^{(l)}(v_i)) + \lambda(1-\sum_{j=1}^{N}q(v_j))\nonumber\\
        =&\sum_{l=1}^{L}\sum_{i=1}^{N}\frac{1}{n_{s_{i}}}\sum_{j=1}^{N}[(\frac{\hat{A}^{2}_{ij}{|h^{(l-1)}(v_j)W^{(l)}|}^{2}}{q(v_j)})\nonumber \\
        -&2\hat{A}_{ij}{h^{(l-1)}(v_j)W^{(l)}}[{F^{(l)}(v_i)}]^T + |F^{(l)}(v_i)|^{2}q(v_j)]\nonumber\\
        +&\lambda(1-\sum_{j=1}^{N}q(v_j))       
    \label{(17)}
\end{align}

\noindent where $n_{s_{i}}$ denotes the number of nodes in subgraph containing node $v_{i}$. However, the node embeddings $h^{(l-1)}(v_j)$ in Eq. (\ref{(17)}) are also related to $q(v_j)$ which leads to the closed-form solution of $q(v_j)$ unable to solve. Critically, it is apparent that $q(v_j)$ can be derived only when the embeddings information are aggregated to calculate the $\mathcal{HW}$ while the structure of subgraph is not defined until the original graph is sampled. Specially, The sampling process is only performed once before training, and the whole training is based on the subgraphs divided by this sampling. To conclude, the circular dependency occurs in subgraph sampling severely.

\begin{figure}[htbp]
    \centering
    \includegraphics[width=3.in]{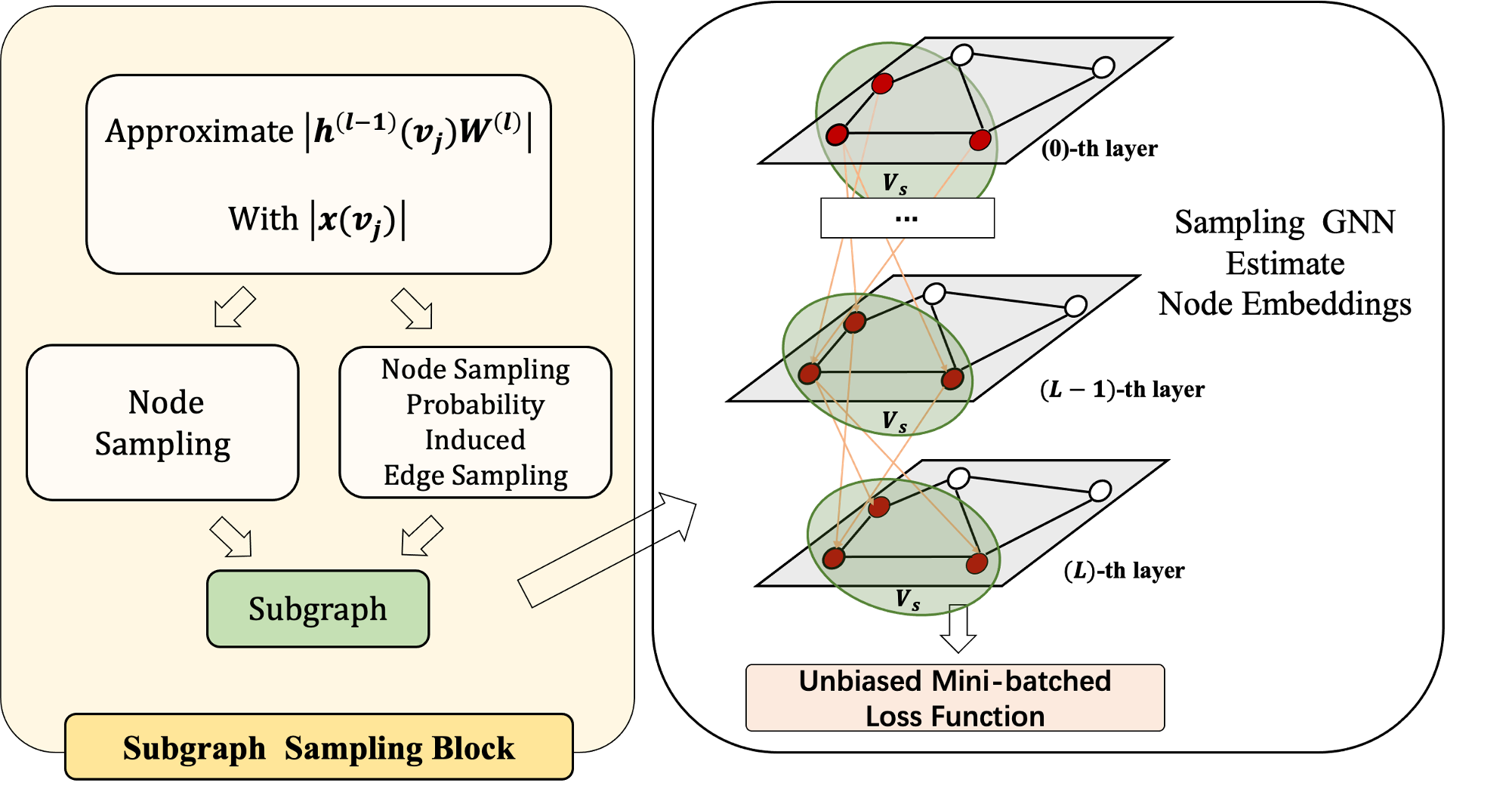} 
    \caption{Overall architecture of HE-SGNN with subgraph sampler.}
    \label{sub}
\end{figure}

\subsubsection{Motivation}  

Circular dependency are more difficult to handle in subgraph sampling. In ClusterGCN, the authors choose to utilize one of cluster algorithm, i.e. METIS \cite{karypis1998fast}, without variance analysis and GraphSAINT also does not make variance analysis for their subgraph node sampler inspired by FastGCN. Additionally, GraphSAINT proposes an unbiased edge sampler from a custom variance analysis framework to outperform the node sampler. However, their edge sampler also takes no information of node embeddings into account for breaking the circular dependency between sampling probability and the $\mathcal{HW}$.

\subsubsection{HE-SGNN with Subgraph Node Sampler} 

Similar to the proposed HE-SGNN in layer-wise sampling, HE-SGNN with Subgraph Node Sampler shown in Fig. \ref{sub} also estimates the $\mathcal{HW}$ firstly and then applies sampling GNN for estimating the node embeddings in vanilla GNN.

\textbf{Estimating the $\mathcal{HW}$.} In subgraph sampling, circular dependency are required to break. Different from layer-wise node sampling, no history information can be utilized for approximating the $\mathcal{HW}$ and subgraph scale $n_{s_{i}}$ in Eq. (\ref{(17)}) iteratively since the subgraph sampling is conducted before the training and the randomly initialized parameters have no useful information. To follow the paradigm of subgraph sampling without any information of $\mathcal{HW}$, we regard the node input features as the embeddings in initial vector space. Therefore, we expect to break the circular dependency by replacing the $\mathcal{HW}$ with the norm of input node features in Eq. (\ref{(17)}). In the meantime, the node sampling probability is expected not to be influenced by the different scales of each subgraph, i. e. $n_{s_{i}}$s,  so we fix the sampled nodes' number $n_{s}$ of each subgraph to substitute different $n_{s_{i}}$s . As a result, Eq. (\ref{(17)}) is approximated as:

\begin{align}
        &L(\lambda,q(v_j))\nonumber\\
        =&\frac{1}{n_{s}}\sum_{l=1}^{L}\sum_{i=1}^{N}\sum_{j=1}^{N}[(\frac{\hat{A}^{2}_{ij}{|x(v_{j})|}^{2}}{q(v_j)})-2\hat{A}_{ij}{x(v_{j})}[{F^{(l)}(v_i)}]^{T}\nonumber \\+& |F^{(l)}(v_i)|^{2}q(v_j)]
        +\lambda(1-\sum_{j=1}^{N}q(v_j))       
    \label{(18)}
\end{align}

Similarly, we can deduce the approximate minimum variance subgraph node sampler by letting 

\begin{equation}
    \begin{split}    
        \frac{\partial L(\lambda,q(v_{j}))}{q(v_{j})} = 0\ \ \ \ s.\ t.\ \frac{\partial L(\lambda,q(v_{j}))}{\lambda} = 0
            \end{split}
    \label{(19)}
\end{equation}

Then we can obtain the approximate minimum variance subgraph node sampler from Eq. (\ref{(19)}) as:

\begin{equation}
    \begin{split}
        {\hat{q}}(v_{j}) = \frac{(\sum_{i=1}^{N}\hat{A}^{2}_{ij})^{\frac{1}{2}}|x(v_{j})|}{\sum_{k=1}^{N}(\sum_{i=1}^{N}\hat{A}^{2}_{ik})^{\frac{1}{2}}|x(v_{k})|}
    \end{split}
    \label{(20)}
\end{equation}

\textbf{Estimating the Node Embeddings for Vanilla GNN.} Finally, given that full batched training loss $Loss=\frac{1}{|V_{t}|}\sum_{i=1}^{|V_{t}|}\mathcal{L}(\hat{y}(v_i), y(v_i))$ where $V_{t}$ denotes the training nodes set and $\mathcal{L}(\hat{y}(v_i), y(v_i))$ denotes the classification loss between predicted label $\hat{y}(v_i)$ for node $v_{i}$ and the ground truth $y(v_i)$, we develop an unbiased subgraph training loss to further eliminate the bias caused by randomness.

\textbf{Theorem 4.} (Proof in Appendix \ref{a4}) \textit{The subgraph training loss ${Loss}(s_{i})$ on $(i)$-th subgraph is an unbiased loss function of full batched training loss $Loss$ as}

\begin{equation}
    \begin{split}
        {Loss}(s_{i}) = \frac{1}{|V_{t}| \cdot n_{s_{i}}}\sum_{j=1}^{n_{s_{i}}}\frac{1}{q(v_{j})}\mathcal{L}(\hat{y}(v_j), y(v_j))
    \end{split}
    \label{(21)}
\end{equation}

To conclude, our proposed HE-SGNN with subgraph node sampler is demonstrated in Fig. \ref{sub}. Firstly, we make approximation to get the importance node sampling probability. Then we partition the original large-scale graph into several subgraphs with this proposed node sampler. Finally, we utilize the unbiased sampling GNN's loss function of full batched training loss to estimating the node embeddings for vanilla GNN as

\begin{equation}
    h^{(l)}(v_i)=\sigma(\frac{1}{n_s}\sum_{j=1}^{|V_s|}\frac{\hat{A}_{ij}}{\hat{q}(v_j)}h^{(l-1)}(v_j)W^{(l)})
    \label{(22)}
\end{equation}

\noindent where $\hat{q}(v_j)$ denotes the approximate sampling probability in subgraph node sampling and $V_s$ denotes the subgraph nodes set.

\begin{table*}[!htp]
\renewcommand\arraystretch{1.75}
\centering
\caption{Dataset Statistics}
\begin{tabular}{p{2.5cm}lp{2.5cm}lp{2.5cm}lp{2.5cm}lp{2.5cm}lp{2.5cm}}
\toprule
\multicolumn{1}{c}{Dataset} & \multicolumn{1}{c}{Nodes} &  \multicolumn{1}{c}{Edges} &  \multicolumn{1}{c}{Degree} & \multicolumn{1}{c}{Features} & \multicolumn{1}{c}{Classes}\\
\midrule
\makecell[c]{Cora} & \makecell[c]{2,708} & \makecell[c]{5,429} & \makecell[c]{2} & \makecell[c]{1,433} &\makecell[c]{7}\\
\makecell[c]{Citeseer} & \makecell[c]{3,327} & \makecell[c]{4,732} & \makecell[c]{1} & \makecell[c]{3,707} &\makecell[c]{6}\\
\makecell[c]{Pubmed} & \makecell[c]{19,717} & \makecell[c]{44,338} & \makecell[c]{2} & \makecell[c]{500} &\makecell[c]{3}\\
\makecell[c]{Flickr} & \makecell[c]{89,250} & \makecell[c]{899,756} & \makecell[c]{10} & \makecell[c]{500} &\makecell[c]{7}\\
\makecell[c]{Reddit} & \makecell[c]{232,965} & \makecell[c]{11,606,919} & \makecell[c]{50} & \makecell[c]{602} &\makecell[c]{41}\\
\makecell[c]{Yelp} & \makecell[c]{716,847} & \makecell[c]{6,977,410} & \makecell[c]{10} & \makecell[c]{300} &\makecell[c]{100}\\
\makecell[c]{Ogbn-products} & \makecell[c]{2,449,029} & \makecell[c]{61,859,140} & \makecell[c]{25} & \makecell[c]{100} &\makecell[c]{47}\\

\bottomrule
\end{tabular}
\label{t1}
\end{table*}

\subsubsection{Node Sampling Probability Induced Edge Sampler} 

Inspired by GraphSAINT, we try to develop an edge sampler and incorporate aggregation information of node into edge samplers to improve sampling process shown in Fig. \ref{sub}. On account of no available historical information, a potential method is to induce edge sampler from the node sampler in Eq. (\ref{(20)}) which leverage input node features to approximate aggregation information, i.e. the $\mathcal{HW}$. For edge sampler, node $v_i$ and $v_j$ are added to a subgraph if $e_{ij}$ is sampled. . So intuitively, the sampling probability of an edge should be proportional to that of its two connected nodes respectively. Additionally, to satisfy the constraint that the sum of sampling probabilities equals to $1$, we induce the edge sampler from node sampling probability as:

\begin{equation}
    \begin{split}
        Pr(e_{ij}) = \frac{q(v_i)}{D(v_i)} + \frac{q(v_j)}{D(v_j)}
    \end{split}
    \label{(23)}
\end{equation}

\noindent where $Pr(e_{ij})$ denotes the sampling probability of the edge connected node $v_i$ and $v_j$. 

\subsubsection{Time Complexity Analysis}

On account that subgraph sampling paradigm conduct graph partition before the entire training process, the difference in efficiency of different subgraph sampling methods is determined by the sampling process including sampling probability calculation and subgraph partition. In particular, our proposed sampler time cost is slightly higher than GraphSAINT due to the external consideration of node input features but still much smaller than the clustering algorithm utilized in ClusterGCN. The actual sampling cost will be shown in detail in the experimental section.



\section{Experiments}

In this section, we evaluate the performance of our proposed HE-SGNN with layer-wise node sampler and two subgraph samplers consisting of the node sampler and the node sampling probability induced edge sampler. Several detailed experiments are conducted to demonstrate the outstanding effectiveness and efficiency of our samplers.

\subsection{Benchmarks and Baselines}

\textbf{Benchmarks.} These experiments are analyzed on different node classification tasks as: 1. classifying the categories of papers in three citation networks (Cora, Citeseer and Pubmed \cite{sen2008collective}), 2. image classification with related descriptions (Flickr), 3. categorizing the communities of posts in one community network (Reddit \cite{hamilton2017inductive}), 4. businesses classification in one community network (Yelp) and 5. classifying the types of products in one recommendation network (Ogbn-products \cite{hu2020open}). The sizes of these seven datasets range from 2.7K (Cora) to 2.4M (Ogbn-products). Specially, the node labels of Yelp are in the multi-class form. The detail of these seven datasets are listed in Table \ref{t1}. To split the datasets for learning, datasets including Cora, Citeseer and Pubmed follow the “fixed-partition” splits in \cite{chen2018fastgcn}, datasets containing Flickr, Reddit and Yelp follow the “fixed-partition” splits in \cite{zeng2019graphsaint} and Ogbn-products follow the “fixed-partition” splits in \cite{hu2020open}.

\textbf{Baselines.} To measure the superiority of our proposed HE-SGNN derived from the unified variance analysis framework, we compare our proposed samplers with three mini-batched representative GNN models including GCN \cite{kipf2016semi}, GraphSAGE \cite{hamilton2017inductive} and GAT \cite{velickovic2017graph} and the other baseline samplers in corresponding sampling paradigms. Specifically, our proposed layer-wise node sampler versus other layer-wise node samplers including FastGCN \cite{chen2018fastgcn} and ASGCN (no variance reduction) \cite{huang2018adaptive}, our proposed subgraph node sampler versus other subgraph node sampler containing ClusterGCN \cite{chiang2019cluster} and GraphSAINT (node sampler) \cite{zeng2019graphsaint} while our proposed subgraph edge sampler versus GraphSAINT (edge sampler) \cite{zeng2019graphsaint}.


\begin{table*}[!htp]
\renewcommand\arraystretch{1.75}
\scriptsize
\centering
\caption{F1-micro Score and Accuracy Comparison with Different Baselines}
\label{t2}
\begin{tabular} {cp{2cm}cp{1.65cm}cp{1.65cm}cp{1.65cm}cp{1.65cm}cp{1.65cm}cp{1.65cm}cp{1.65cm}cp{1.65cm}}
\toprule
\multicolumn{1}{c}{Paradigm} &\multicolumn{1}{c}{Method} & \multicolumn{1}{c}{Cora} &  \multicolumn{1}{c}{Citeseer} & \multicolumn{1}{c}{Pubmed} &  \multicolumn{1}{c}{Flickr}& \multicolumn{1}{c}{Reddit}& \multicolumn{1}{c}{Yelp}& \multicolumn{1}{c}{Ogbn-products}\\
\midrule

\multirow{3}{*}{\makecell[c]{Mini\\Batched}} & \makecell[c]{GCN} & \makecell[c]{0.851$\pm$ 0.001} & \makecell[c]{0.770$\pm$ 0.003} & \makecell[c]{0.867$\pm$ 0.001}& \makecell[c]{0.492 $\pm$ 0.003} & \makecell[c]{0.933 $\pm$ 0.000}&  \makecell[c]{0.378$\pm$ 0.001}&  \makecell[c]{0.758$\pm$ 0.002}\\ 
&\makecell[c]{GraphSAGE} & \makecell[c]{0.822$\pm$ 0.010} & \makecell[c]{0.714$\pm$ 0.008} & \makecell[c]{0.873$\pm$ 0.002}& \makecell[c]{0.501$\pm$ 0.010}&\makecell[c]{0.953$\pm$ 0.001}& \makecell[c]{0.634$\pm$ 0.010}&\makecell[c]{0.782$\pm$ 0.010}\\ 
&\makecell[c]{GAT} & \makecell[c]{0.863$\pm$ 0.001} & \makecell[c]{0.769$\pm$ 0.003} & \makecell[c]{0.870$\pm$ 0.001}                  & \makecell[c]{0.507$\pm$ 0.002}&\makecell[c]{---}& \makecell[c]{---}&\makecell[c]{---}\\
\midrule
\multirow{3}{*}{\makecell[c]{Layer-wise\\Sampling}} & \makecell[c]{FastGCN} & \makecell[c]{0.850$\pm$ 0.005} & \makecell[c]{0.776$\pm$ 0.008} & \makecell[c]{0.880$\pm$ 0.010}& \makecell[c]{0.504$\pm$ 0.001} & \makecell[c]{0.922$\pm$ 0.001}&  \makecell[c]{0.261$\pm$ 0.050}&  \makecell[c]{0.316$\pm$ 0.167}\\ 
&\makecell[c]{ASGCN} & \makecell[c]{0.861$\pm$ 0.006} & \makecell[c]{\textbf{0.792$\pm$ 0.008}} & \makecell[c]{\textbf{0.894 $\pm$ 0.002}}& \makecell[c]{0.504 $\pm$ 0.002}&\makecell[c]{\textbf{0.945 $\pm$ 0.005}}& \makecell[c]{---}&\makecell[c]{---}\\ 
&\makecell[c]{\textbf{HE-SGNN(L)}} & \makecell[c]{\textbf{0.872 $\pm$ 0.001}} & \makecell[c]{0.789$\pm$ 0.004} & \makecell[c]{\textbf{0.894$\pm$ 0.003}}                  & \makecell[c]{\textbf{0.507$\pm$ 0.002}}&\makecell[c]{0.938$\pm$ 0.001}& \makecell[c]{\textbf{0.390$\pm$ 0.017}}&\makecell[c]{\textbf{0.683$\pm$ 0.016}}\\
\midrule
\multirow{5}{*}{\makecell[c]{Subgraph\\Sampling}} & \makecell[c]{ClusterGCN} & \makecell[c]{0.847$\pm$ 0.005} & \makecell[c]{0.717$\pm$ 0.012} & \makecell[c]{0.884$\pm$ 0.005}& \makecell[c]{0.481$\pm$ 0.004} & \makecell[c]{0.954$\pm$ 0.001}&  \makecell[c]{0.609$\pm$ 0.005}&  \makecell[c]{0.790$\pm$ 0.003}\\ 
&\makecell[c]{GraphSAINT(SN)} & \makecell[c]{0.851$\pm$ 0.002} & \makecell[c]{0.766$\pm$ 0.007} & \makecell[c]{\textbf{0.898 $\pm$ 0.001}}& \makecell[c]{0.507$\pm$ 0.003}&\makecell[c]{0.966$\pm$ 0.001}& \makecell[c]{0.641$\pm$ 0.000}&\makecell[c]{0.788$\pm$ 0.002}\\ 
&\makecell[c]{\textbf{HE-SGNN(SN)}} & \makecell[c]{\textbf{0.860$\pm$ 0.003}} & \makecell[c]{\textbf{0.777$\pm$ 0.007}} & \makecell[c]{\textbf{0.902$\pm$ 0.002}}                  & \makecell[c]{\textbf{0.511$\pm$ 0.003}}&\makecell[c]{\textbf{0.970$\pm$ 0.001}}& \makecell[c]{\textbf{0.642$\pm$ 0.000}}&\makecell[c]{\textbf{0.794$\pm$ 0.001}}\\
\cline{2-9}
&\makecell[c]{GraphSAINT(SE)} & \makecell[c]{0.856$\pm$ 0.003} & \makecell[c]{0.753$\pm$ 0.011} & \makecell[c]{\textbf{0.900$\pm$ 0.002}}& \makecell[c]{0.510$\pm$ 0.002}&\makecell[c]{0.966$\pm$ 0.001}& \makecell[c]{\textbf{0.653$\pm$ 0.000}}&\makecell[c]{0.799$\pm$ 0.001}\\ 
&\makecell[c]{\textbf{HE-SGNN(SE)}} & \makecell[c]{\textbf{0.863$\pm$ 0.001}} & \makecell[c]{\textbf{0.784$\pm$ 0.003}} & \makecell[c]{0.896$\pm$ 0.001}                  & \makecell[c]{\textbf{0.517$\pm$ 0.003}}&\makecell[c]{\textbf{0.970$\pm$ 0.000}}& \makecell[c]{0.649$\pm$ 0.000}&\makecell[c]{\textbf{0.803$\pm$ 0.001}}\\
\bottomrule
\end{tabular}
\end{table*}

\begin{table}[!htp]
\renewcommand\arraystretch{1.75}
\scriptsize
\centering
\caption{More Detailed Accuracy Comparison on Reddit and Ogbn-products}
\begin{tabular}{cp{2cm}cp{2cm}cp{1.8cm}cp{1.8cm}}
\toprule
\multicolumn{1}{c}{Paradigm} &\multicolumn{1}{c}{Method} & \multicolumn{1}{c}{Reddit} &  \multicolumn{1}{c}{Ogbn-products}\\
\midrule

\multirow{6}{*}{\makecell*[b]{Mini\\Batched}} & \makecell[c]{GCN} & \makecell[c]{0.933 $\pm$ 0.000}&    \makecell[c]{0.758$\pm$ 0.002}\\ 
&\makecell[c]{GraphSAGE}&\makecell[c]{0.953$\pm$ 0.001}& \makecell[c]{0.782$\pm$ 0.010}\\ 
&\makecell[c]{GAT} & \makecell[c]{---}&\makecell[c]{---}\\
&\makecell[c]{SIGN} & \makecell[c]{0.968$\pm$ 0.001}&\makecell[c]{0.776$\pm$ 0.015}\\
&\makecell[c]{GASGCN} &\makecell[c]{0.955 $\pm$ 0.004}&\makecell[c]{0.766 $\pm$ 0.021}\\
&\makecell[c]{SHADOWGCN}  & \makecell[c]{0.958$\pm$ 0.005}&\makecell[c]{0.774$\pm$ 0.005}\\
\midrule
\multirow{4}{*}{\makecell[b]{Layer-wise\\Sampling}} & \makecell[c]{FastGCN} & \makecell[c]{0.924$\pm$ 0.001}&   \makecell[c]{0.316$\pm$ 0.067}\\
&\makecell[c]{LADIES} &\makecell[c]{0.928 $\pm$ 0.002}&\makecell[c]{0.652 $\pm$ 0.034}\\
&\makecell[c]{ASGCN} &\makecell[c]{\textbf{0.951 $\pm$ 0.005}}&\makecell[c]{---}\\ 
&\makecell[c]{\textbf{HE-SGNN(L)}}&\makecell[c]{0.943$\pm$ 0.001}&\makecell[c]{\textbf{0.683$\pm$ 0.016}}\\
\midrule
\multirow{5}{*}{\makecell[b]{Subgraph\\Sampling}} & \makecell[c]{ClusterGCN} & \makecell[c]{0.954$\pm$ 0.001}&   \makecell[c]{0.790$\pm$ 0.003}\\ 
&\makecell[c]{GraphSAINT(SN)}&\makecell[c]{0.966$\pm$ 0.001}&\makecell[c]{0.788$\pm$ 0.002}\\ 
&\makecell[c]{\textbf{HE-SGNN(SN)}} &\makecell[c]{\textbf{0.970$\pm$ 0.001}}&\makecell[c]{\textbf{0.794$\pm$ 0.001}}\\
\cline{2-4}
&\makecell[c]{GraphSAINT(SE)} &\makecell[c]{0.966$\pm$ 0.001}&\makecell[c]{0.799$\pm$ 0.001}\\ 
&\makecell[c]{\textbf{HE-SGNN(SE)}} &\makecell[c]{\textbf{0.970$\pm$ 0.000}}&\makecell[c]{\textbf{0.803$\pm$ 0.001}}\\
\bottomrule 
\end{tabular}
\label{t3}
\end{table}

\subsection{Experimental Settings and Configurations}

\textbf{Settings.} In order to be consistent with the baselines, our proposed hierarchical estimation based models also conduct the experiments in the inductive way. At the meantime, optimized sampler is only applied for training and all neighbours of one node are used for embeddings aggregation in test process. All the baselines and our proposed samplers are implemented in Pytorch version with CUDA 10.2 and on a NVIDIA Tesla V100 GPU with 32GB GPU memory. "F1-micro" score is utilzed for measure the multilabel classification performance of baselines and our proposed models and "F1-micro" score is equivalent to "Accuracy" for single-label node classification. 

\textbf{Configurations.} Adam optimizer \cite{kingad2015methodforstochasticoptimization} is employed for training our proposed HE-SGNN and all baselines. Then we conduct grid search on various hyperparameter configurations to select the configurations with the highest average "F1-micro" score of 10 trials on the validation set as the best model configurations for our HE-SGNN and all the baselines. The best configurations of our HW-SGNN are listed in Appendix \ref{b} with Table \ref{t4}. 


\subsection{Node Classification Performance Analysis}

Table \ref{t2} demonstrates the F1-micro score and accuracy comparison of our proposed models and other baselines. In Table \ref{t2}, "L" denotes layer-wise node sampler, "SN" denotes subgraph node sampler while "SE" denotes subgraph edge sampler. Horizontally, our proposed layer-wise node sampler, subgraph node sampler and subgraph edge sampler outperform other baselines in corresponding sampling paradigms respectively. Vertically, layer-wise sampling based GNNs have higher "F1-micro" scores on smaller and sparser datasets such as Cora and Citeseer than traditional mini-batched GNNs and subgraph sampling based GNNs. However, layer-wise sampling based GNNs lose competitiveness on larger and denser graphs while subgraph sampling based GNNs show superb performance than other algorithms. It is most probably due to the reason that the resampling before every epoch endow the GNN with the ability to aggregate each neighbour's information on sparse graphs while the increase in the number of neighbors causes a large number of neighbors to be discarded when aggregating the information and the variance is still difficult to eliminate in practice due to the randomness of sampling. 

Meanwhile, in order to make the scalability of our proposed hierarchical estimation based models more convincing, one impressive variant of GCN: SGC \cite{wu2019simplifying}, two scalable GNN algorithm: GAS \cite{fey2021gnnautoscale} and SHADOW \cite{zeng2021decoupling}, another layer-wise sampling based GCN: LADIES, are added to highlight the representation learning performance of our proposed HE-SGNN on two large graph datasets including Reddit and Ogbn-products in Table \ref{t3}. Apparently, our proposed three sampler based HE-SGNN still have excellent performance compared to other representative models in corresponding sampling paradigm respectively. Particularly, our proposed HE-SGNN with subgraph edge sampler reaches the best performance on both Reddit and Ogbn-products.

\begin{figure}[htbp]
    \centering
    \subfloat[]{
        \label{fig5}
        \includegraphics[width=2.8in,height=0.15\textheight]{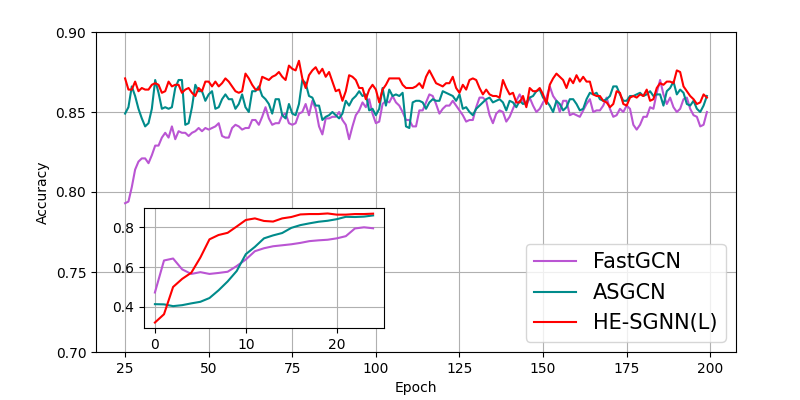}}
    \hspace{0.0pt} 
    \subfloat[]{
        \label{fig6}
        \includegraphics[width=2.8in,height=0.15\textheight]{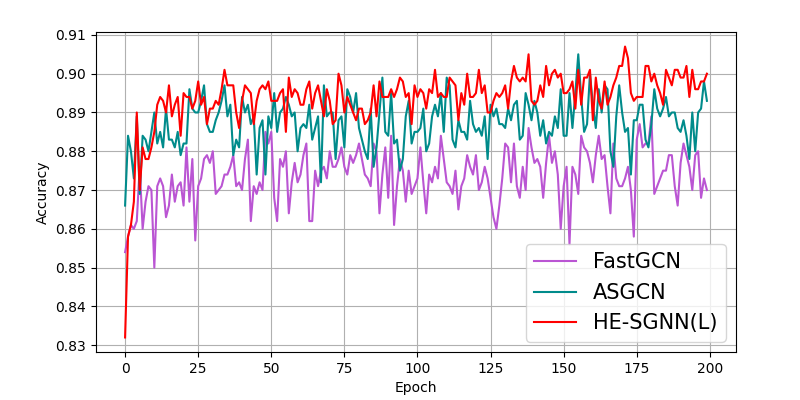} }
    \caption{(a) Test accuracy curves of layer-wise sampling on Cora. (b) Test accuracy curves of layer-wise sampling on Pubmed.}
    \label{fig3}
\end{figure}

\subsection{Running Time Analysis}

\textbf{Layer-wise Sampling.} We compare the training time per epoch of our model with FastGCN, ASGCN experimentally. In fairness to all model, layer numbers, sampling size and dimensions in each layer of the obove three models are consistent with the configurations of our proposed model. Fig. \ref{fig3} demonstrates the comparison of training time per epoch on Pubmed and Reddit. The results confirm that the time per epoch of our model is slightly higher than that of FastGCN and much lower than that of ASGCN. Considering the effectiveness as well, our model has a large improvement in accuracy while suffers a little increase of time cost.

\begin{figure}[htbp]
    \centering
    \includegraphics[width=2.in,height=0.2\textheight]{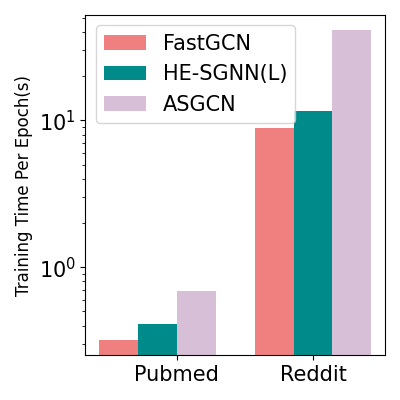} 
    \caption{Running time comparison.}
    \label{fig3}
\end{figure}

\begin{table*}[!htp]
\renewcommand\arraystretch{1.75}
\scriptsize
\centering
\caption{Sampling Cost (seconds) Comparison of Subgraph Sampling Methods}
\begin{tabular}{cp{1.5cm}cp{1.5cm}cp{1.5cm}cp{1.5cm}cp{1.5cm}cp{1.5cm}cp{1.5cm}cp{1.5cm}cp{1.5cm}}
\toprule
\multicolumn{1}{c}{Method} & \multicolumn{1}{c}{Cora} &  \multicolumn{1}{c}{Citeseer} & \multicolumn{1}{c}{Pubmed} &  \multicolumn{1}{c}{Flickr}& \multicolumn{1}{c}{Reddit}& \multicolumn{1}{c}{Yelp}& \multicolumn{1}{c}{Ogbn-products}\\
\midrule
\makecell[c]{ClusterGCN} & \makecell[c]{$0.072$} & \makecell[c]{$0.159$}  & \makecell[c]{$0.977$} & \makecell[c]{$8.321$}& \makecell[c]{$28.371$}& \makecell[c]{$70.262$}& \makecell[c]{$2172.627$}\\ 
\makecell[c]{GraphSAINT(SN)} & \makecell[c]{$0.024$} & \makecell[c]{$0.075$}  & \makecell[c]{$0.294$} & \makecell[c]{$0.452$}& \makecell[c]{$6.946$}& \makecell[c]{$19.780$}& \makecell[c]{$4.348$}\\
\makecell[c]{\textbf{HE-SGNN(SN)}} & \makecell[c]{$0.015$} & \makecell[c]{$0.072$}  & \makecell[c]{$0.156$} & \makecell[c]{$0.447$}& \makecell[c]{$2.275$}& \makecell[c]{$4.791$}& \makecell[c]{$1.424$}\\
\makecell[c]{GraphSAINT(SE)} & \makecell[c]{$0.027$} & \makecell[c]{$0.074$}  & \makecell[c]{$0.306$} & \makecell[c]{$0.580$}& \makecell[c]{$13.361$}& \makecell[c]{$29.837$}& \makecell[c]{$23.132$}\\
\makecell[c]{\textbf{HE-SGNN(SE)}} & \makecell[c]{$0.031$} & \makecell[c]{$0.078$}  & \makecell[c]{$0.327$} & \makecell[c]{$0.873$}& \makecell[c]{$24.263$}& \makecell[c]{$25.035$}& \makecell[c]{$27.061$}\\ 
\bottomrule
\end{tabular}

\label{t6}
\end{table*}

\textbf{Subgraph Sampling.} As a result, we demonstrate the different time costs on sampling process for subgraph sampling baselines with their optimal configurations in Table \ref{t6}. Table \ref{t6} confirms the cost of our proposed samplers are still in the same order of magnitude as the sampling cost of GraphSAINT. Meanwhile, the clustering algorithm utilized in ClusterGCN shows expensive sampling cost when the graph scale is large. Overall, the sampling costs of our proposed subgraph node sampler and edge sampler are still small compared with other baselines.

\begin{figure}[htbp]
    \centering
    \subfloat[]{
        \label{Sub11}
        \includegraphics[width=2.4in,height=0.15\textheight]{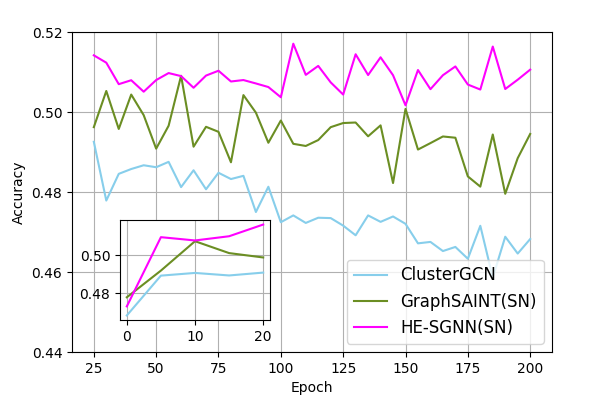}
    }
    \hspace{0.0pt} 
    \subfloat[]{
        \label{Sub12}
        \includegraphics[width=2.4in,height=0.15\textheight]{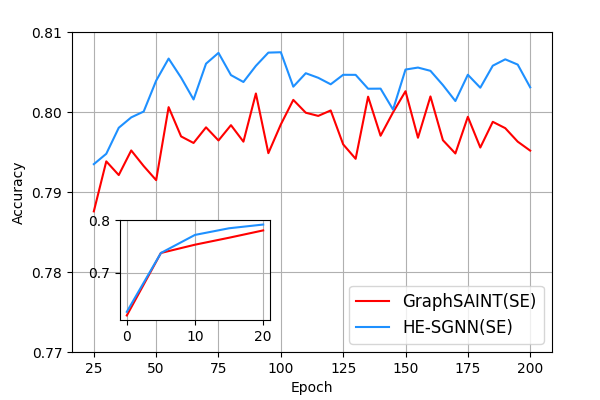} 
    }
    \caption{(a) Test accuracy curves of subgraph node sampling on Flickr. (b) Test accuracy curves of subgraph edge sampling on Ogbn-products.}
    \label{Sub1}
\end{figure}

\subsection{Variance Analysis}

The convergence depends on the variance introduced by sampling significantly. Therefore, the test accuracy curves of our models and other baselines are demonstrated to prove the effectiveness of our proposed three samplers.

\textbf{Test Accuracy Curves of Layer-wise Sampling.} The accuracy curves of FastGCN, ASGCN and our proposed HE-SGNN(L) on Cora and Pubmed are drawn in Fig. \ref{fig3}. Due to the encouragement of exploration with optimistic initialization, test accuracies of our sampling based model in the first few epochs are lower that those of FastGCN and ASGCN. However, the performance of our hierarchical estimation based sampling GNN improves rapidly after fully exploration. Results shown in \ref{fig3} verify that our HE-SGNN(L) has a faster convergence speed in Cora and almost converges synchronously with ASGCN on Pubmed.

\textbf{Test Accuracy Curves of Subgraph Sampling.} Fig. \ref{Sub11} shows the accuracy curves of ClusterGCN, GraphSAINT(SN) and HE-SGNN(SN) while Fig. \ref{Sub12} demonstrates the convergence performance of GraphSAINT(SE) and HE-SGNN(SE). Our proposed HE-SGNN(SN) and HE-SGNN(SE) also converge quickly. Overall, HE-SGNN(SN) and our HE-SGNN(SE) also have superior convergence ability, inferring the effectiveness in reducing variance of proposed two subgraph samplers.

\subsection{Ablation Study}

\textbf{Initialization Analysis for Layer-wise Node Sampler.} As mentioned above, the $\mathcal{HW}$ of each node in each layer are expected to be explored for precise valuation which provides guarantee for the validity of optimistic initialization.

Therefore, in order to demonstrate the importance of optimistic initialization of Eq. (\ref{(17)}), we compare the performance of our proposed layer-wise node sampler with different initialization strategies on Cora and Citeseer in Fig. \ref{fig11}. We set $E^{(0)}(\mathcal{HW}_{v_{j}})$ of each node $v_j$ to $1000$ for optimistic initialization while we set $E^{(0)}(\mathcal{HW}_{v_{j}})$ of each node $v_j$ to $1$ for pessimistic initialization. From Fig. \ref{fig11} we can find that the initial encouragement of exploration let the model gain considerable learning ability and converge quickly with more reasonable exploitation after showing poor performance on account of exploration in the beginning.

\begin{figure}[htbp]
    \centering
    \subfloat[]{
        \label{fig9}
        \includegraphics[width=2.75in,height=0.151\textheight]{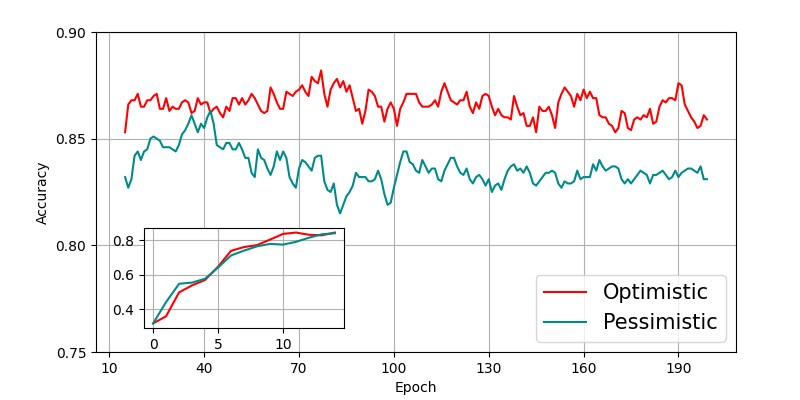}
    }
    \hspace{0.0pt} 
    \subfloat[]{
        \label{fig10}
        \includegraphics[width=2.75in,height=0.151\textheight]{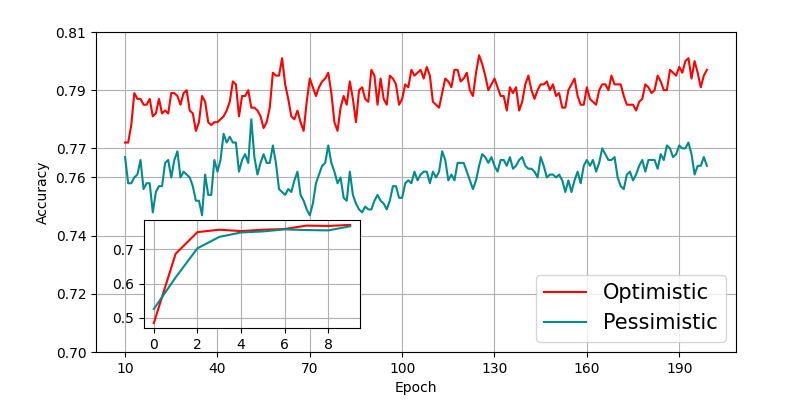} 
    }
    \caption{(a) Test accuracy curves of layer-wise sampler with different initialization strategies on Cora. (b) Test accuracy curves of layer-wise sampler with different initialization strategies on Citeseer.}
    \label{fig11}
\end{figure}

\textbf{Sampling Difference between HE-SGNN(SN) and HE-SGNN(SE).} Additionally, The difference in performance between HE-SGNN(SN) and HE-SGNN(SE) is also worth pondering. Fig. \ref{R11} compares the F1-micro score / test accuracy curves between HE-SGNN(SN) and HE-SGNN(SE) on Reddit and Yelp. It is apparent that these two models both performs well on Reddit while the HE-SGNN(SE) outperforms our HE-SGNN(SN) on Yelp. Potentially, the edge sampler guarantee that each sampled node has at least one neighbour for message passing while the node sampler probably samples several isolated nodes lacking useful aggregation information when the graph is much sparse. Therefore, on Reddit dataset of which average degree is large enough, these two models both converges well while the HE-SGNN(SN) encounters bottleneck on sparse dataset Yelp.

\begin{figure}[htbp]
    \centering
    \subfloat[]{
        \label{R11}
        \includegraphics[width=2.4in,height=0.15\textheight]{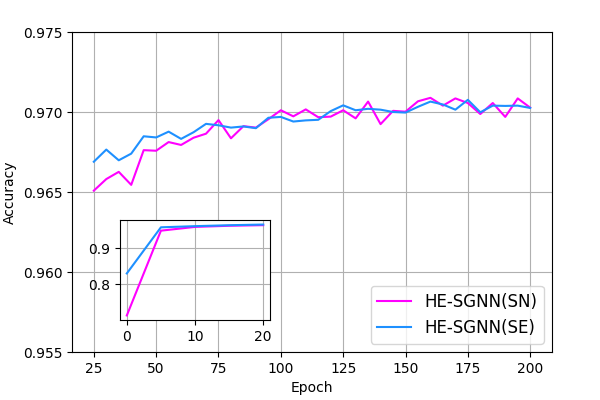}
    }
    \hspace{0.0pt} 
    \subfloat[]{
        \label{R12}
        \includegraphics[width=2.4in,height=0.15\textheight]{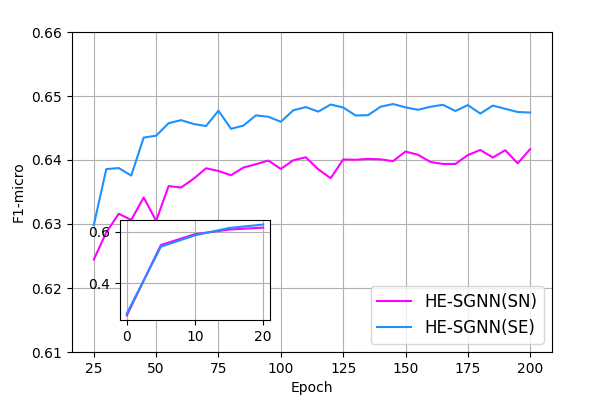} 
    }
    \caption{(a) Accuracy curves comparison between HE-SGNN(SN) and HE-SGNN(SE) on Reddit. (b) F1-micro score curves comparison between HE-SGNN(SN) and HE-SGNN(SE) on Yelp.}
    \label{R1}
\end{figure}



\section{Conclusion}

To improve the scalability of GCN, we define an unified variance analysis framework of existing three node sampling paradigms (node-wise sampling, layer-wise sampling and subgraph sampling) into one framework and point out that critical challenges for variance reduction, i.e. circular dependency. Considering the remaining exponential explosion of neighbours' scale in node-wise sampling, we focus on the sampler optimization for layer-wise sampling and subgraph sampling. To combat the circular dependency in layer-wise sampling, we propose a hierarchical estimation based sampling GNN (HE-SGNN). In detail, we utilize the historical information which can be regarded as a time series sampling to estimate current unknown $\mathcal{HW}$ with optimistic initialization, thus breaking the circular dependency and then employing node sampling to sample nodes for estimating the node embeddings for vanilla GNN. To reduce variance in subgraph sampling, we break the circular dependency to obtain the node sampler by estimating the current unknown $\mathcal{HW}$ with the input node features, from which we induce an edge sampler to further improve the sampling process. The experimental results on seven prevalent graph datasets demonstrate the effectiveness and efficiency of our proposed three samplers by the comparison with existing SOTA methods.

\section*{Acknowledgments}
This work was supported by the National Natural Science Foundation of China (Grant No.6220072896 and No.U21B2046 ). and Postdoctoral Science Fund (No.292151).

\ifCLASSOPTIONcaptionsoff
  \newpage
\fi



%



{
\small
\bibliographystyle{IEEEtran}
\bibliography{ref}
}

%

\begin{IEEEbiography}[{\includegraphics[width=1in,height=1.25in,clip,keepaspectratio]{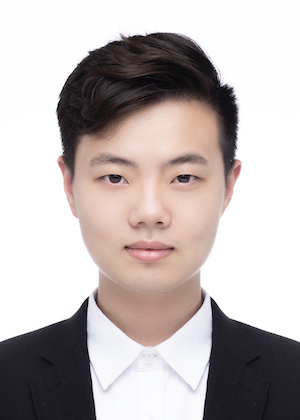}}]{Yang Li} is currently a MS student in the Data Intelligence System Research Center, Institute of Computing Technology, Chinese Academy of Sciences (CAS). He received his bachelor degree from Nanjing University of Information Science and Technology in 2021. His research interests include scalable graph neural networks and privacy attack on graph neural networks. 
\end{IEEEbiography}

\begin{IEEEbiography}[{\includegraphics[width=1in,height=1.25in,clip,keepaspectratio]{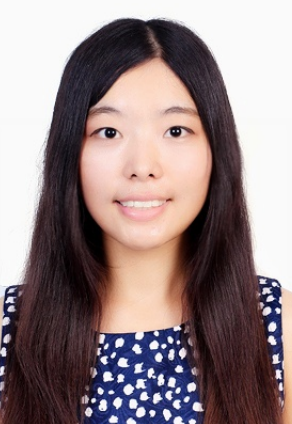}}]{Bingbing Xu}
is an assistant professor at Institute of Computing Technology, Chinese Academy of Sciences (CAS). She received her Ph.D. degree from Institute of Computing Technology, CAS in 2021. Her research interests include geometric deep learning, graph based data mining, graph convolution and trustworthy artificial intelligence.
\end{IEEEbiography}


\begin{IEEEbiography}[{\includegraphics[width=0.9in,height=1.25in,clip,keepaspectratio]{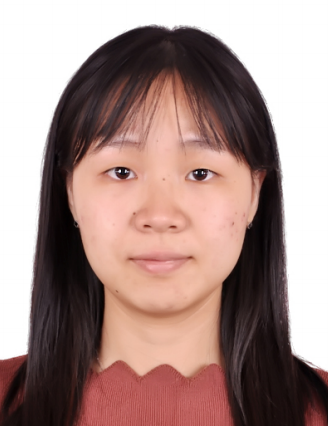}}]{Qi Cao}
is an assistant professor at Institute of Computing Technology, Chinese Academy of Sciences (CAS). She received her Ph.D. degree from Institute of Computing Technology, CAS in 2020. Her research interests include social media computing, influence modeling, information diffusion prediction and adversarial attack and defense on raph neural networks.
\end{IEEEbiography}

\begin{IEEEbiography}[{\includegraphics[width=1in,height=1.25in,clip,keepaspectratio]{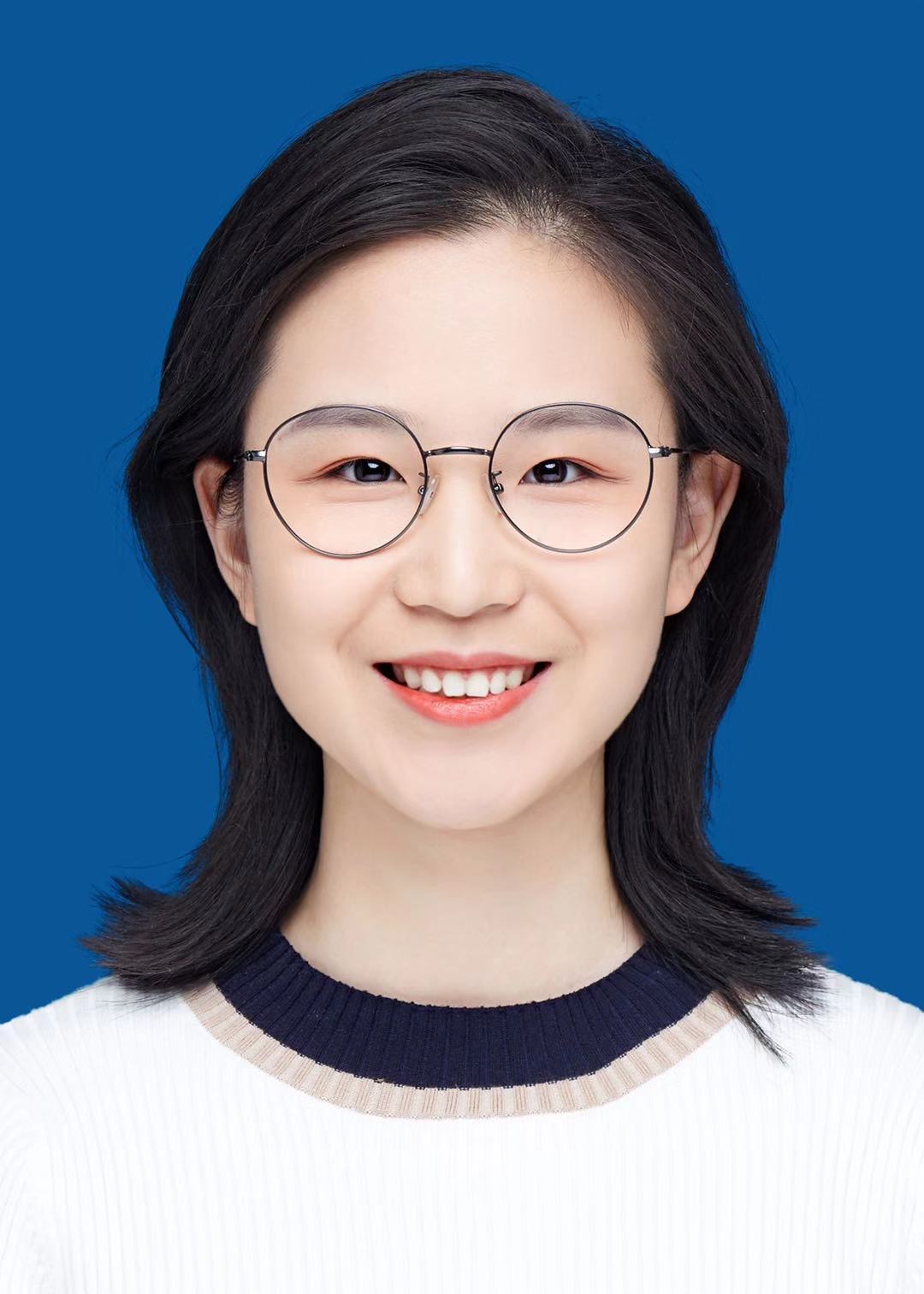}}]{Yige Yuan} is currently a Ph.D. student in the Data Intelligence System Research Center, Institute of Computing Technology, Chinese Academy of Sciences (CAS). She received her bachelor degree from Xidian University in 2020. Her research interests include generalization in graph representation learning and trustworthy artificial intelligence. 
\end{IEEEbiography}

\begin{IEEEbiography}[{\includegraphics[width=0.9in,height=1.25in,clip,keepaspectratio]{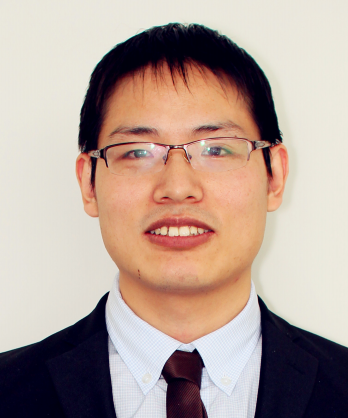}}]{Huawei Shen}
is a professor in the Institute of Computing Technology, Chinese Academy of Sciences (CAS) and the University of Chinese Academy of Sciences. He received his Ph.D. degree from the Institute of Computing Technology in 2010. His major research interests include network science, social media analytics and recommendation. He has published more than 140 papers in prestigious journals and top international conferences, including in Science, PNAS, Physical Review E, WWW, AAAI, IJCAI, SIGIR, CIKM, and WSDM. He is an Outstanding Member of the Association of Innovation Promotion for Youth of CAS. He received the Top 100 Doctoral Thesis Award of CAS in 2011 and the Grand Scholarship of the President of CAS in 2010.
\end{IEEEbiography}




\clearpage
\appendices

\section{Related Proof}
\subsection{Proof of Theorem 1}
\label{a1}
$\hat{F}^{(l)}(v_i)$ is an unbiased estimator of $F^{(l)}(v_i)$ when $\mathbb{E}_{q_{j}^{(l-1)}}(\hat{F}^{(l)}(v_i)) = F^{(l)}(v_i)$. So the proof is decomposed as follows:

\begin{align}
        &\mathbb{E}_{q_{j}^{(l-1)}}(\hat{F}^{(l)}(v_i))\nonumber\\
        =&\mathbb{E}_{q_{j}^{(l-1)}}(\frac{1}{n_{i}^{(l-1)}}\sum_{j \in V_i^{(l-1)}}\frac{p(v_{j}|v_{i})}{q_{j}^{(l-1)}}h^{(l-1)}(v_j)W^{(l)})\nonumber\\
        =&\mathbb{E}_{q_{j}^{(l-1)}}(\frac{p(v_{j}|v_{i})}{q_{j}^{(l-1)}}h^{(l-1)}(v_j)W^{(l)})\nonumber\\
        =&\sum_{j=1}^{N}p(v_{j}|v_{i})h^{(l-1)}(v_j)W^{(l)}\nonumber\\
        =&F^{(l)}(v_i)
    \label{(24)}
\end{align}

\begin{table*}[!htp]
\renewcommand\arraystretch{1.75}
\scriptsize
\centering
\caption{Training Configurations of Our Proposed Models}
\begin{tabular}{cp{1.5cm}cp{1.5cm}cp{1.5cm}cp{1.5cm}cp{1.5cm}cp{1.5cm}cp{1.5cm}cp{1.5cm}cp{1.5cm}cp{1.5cm}}
\toprule
\multicolumn{1}{c}{Paradigm} &\multicolumn{1}{c}{Method} &\multicolumn{1}{c}{Hyperparameter} & \multicolumn{1}{c}{Cora} &  \multicolumn{1}{c}{Citeseer} & \multicolumn{1}{c}{Pubmed} &  \multicolumn{1}{c}{Flickr}& \multicolumn{1}{c}{Reddit}& \multicolumn{1}{c}{Yelp}& \multicolumn{1}{c}{Ogbn-products}\\
\midrule
\multirow{4}{*}{\makecell[c]{Layer-wise\\Sampling}} 
& \multirow{4}{*}{\makecell[c]{HE-SGNN(L)}} & \makecell[c]{Dropout} & \makecell[c]{$0.0$} & \makecell[c]{$0.0$}  & \makecell[c]{$0.0$} & \makecell[c]{$0.4$}& \makecell[c]{$0.0$}& \makecell[c]{$0.0$}& \makecell[c]{$0.0$}\\ 
&&\makecell[c]{Batch size}& \makecell[c]{$256$}& \makecell[c]{$256$}&\makecell[c]{$256$}&\makecell[c]{$512$}&\makecell[c]{$512$}&\makecell[c]{$512$}&\makecell[c]{$512$}\\
&&\makecell[c]{Hidden dimension}& \makecell[c]{$16$}  & \makecell[c]{$16$} &  \makecell[c]{$16$} & \makecell[c]{$256$}& \makecell[c]{$16$}& \makecell[c]{$512$}& \makecell[c]{$256$}\\
&&\makecell[c]{Activation function}& \makecell[c]{Sigmoid}  & \makecell[c]{Sigmoid} &  \makecell[c]{Sigmoid} & \makecell[c]{ReLU}& \makecell[c]{Sigmoid}& \makecell[c]{ReLU}& \makecell[c]{ReLU}\\
\midrule
\multirow{8}{*}{\makecell[c]{Subgraph\\Sampling}} 
& \multirow{4}{*}{\makecell[c]{HE-SGNN(SN)}} & \makecell[c]{Dropout} & \makecell[c]{$0.5$} & \makecell[c]{$0.4$}  & \makecell[c]{$0.2$} & \makecell[c]{$0.5$}& \makecell[c]{$0.2$}& \makecell[c]{$0.1$}& \makecell[c]{$0.3$}\\
&&\makecell[c]{Batch size}& \makecell[c]{$512$}& \makecell[c]{$512$}&\makecell[c]{$5000$}&\makecell[c]{$10000$}&\makecell[c]{$20000$}&\makecell[c]{$20000$}&\makecell[c]{$20000$}\\
&&\makecell[c]{Hidden dimension}& \makecell[c]{$16$}& \makecell[c]{$16$}&\makecell[c]{$16$}&\makecell[c]{$256$}&\makecell[c]{$256$}&\makecell[c]{$512$}&\makecell[c]{$256$}\\
&&\makecell[c]{Activation function}& \makecell[c]{ReLU}  & \makecell[c]{ReLU} &  \makecell[c]{ReLU} & \makecell[c]{ReLU}& \makecell[c]{ReLU}& \makecell[c]{ReLU}& \makecell[c]{ReLU}\\
\cline{2-10}
& \multirow{4}{*}{\makecell[c]{HE-SGNN(SE)}} & \makecell[c]{Dropout} & \makecell[c]{$0.5$} & \makecell[c]{$0.5$}  & \makecell[c]{$0.2$} & \makecell[c]{$0.4$}& \makecell[c]{$0.3$}& \makecell[c]{$0.1$}& \makecell[c]{$0.3$}\\
&&\makecell[c]{Batch size}& \makecell[c]{$512$}& \makecell[c]{$512$}&\makecell[c]{$5000$}&\makecell[c]{$20000$}&\makecell[c]{$10000$}&\makecell[c]{$5000$}&\makecell[c]{$20000$}\\
&&\makecell[c]{Hidden dimension}& \makecell[c]{$16$}& \makecell[c]{$16$}&\makecell[c]{$16$}&\makecell[c]{$256$}&\makecell[c]{$256$}&\makecell[c]{$512$}&\makecell[c]{$256$}\\
&&\makecell[c]{Activation function}& \makecell[c]{ReLU}  & \makecell[c]{ReLU} &  \makecell[c]{ReLU} & \makecell[c]{ReLU}& \makecell[c]{ReLU}& \makecell[c]{ReLU}& \makecell[c]{ReLU}\\
\bottomrule
\end{tabular}
\label{t4}
\end{table*}

\subsection{Proof of Theorem 2}
\label{a2}
Given that the unbiased estimation $ \hat{F}^{(l)}(v_i)$, the variance of $\hat{F}^{(l)}(v_i)$ can be derived as

\begin{align}
\small
        &Var_{q^{(l-1)}_{j}}(\hat{F}^{(l-1)}({v_i})))\nonumber\\
        =&Var_{q^{(l-1)}_{j}}(\frac{1}{n_{i}^{(l-1)}}\sum_{j \in V_i^{(l-1)}}\frac{p(v_{j}|v_{i})}{q_{j}^{(l-1)}}h^{(l-1)}(v_j)W^{(l)})\nonumber\\
        =&(\frac{1}{n_{i}^{(l-1)}})^{2}Var_{q^{(l-1)}_{j}}(\sum_{j \in V_i^{(l-1)}}\frac{p(v_{j}|v_{i})}{q_{j}^{(l-1)}}h^{(l-1)}(v_j)W^{(l)})\nonumber\\
        =&\frac{1}{n_{i}^{(l-1)}}Var_{q^{(l-1)}_{j}}(\frac{p(v_{j}|v_{i})}{q_{j}^{(l-1)}}h^{(l-1)}(v_j)W^{(l)})\nonumber\\
        =&\frac{1}{n_{i}^{(l-1)}}\mathbb{E}_{q^{(l-1)}_{j}}[(\frac{p(v_{j}|v_{i})}{q_{j}^{(l-1)}}h^{(l-1)}(v_j)W^{(l)}-{F}^{(l)}(v_i))^{2}]\nonumber\\
        =&\mathbb{E}_{q_{j}^{(l-1)}}[\frac{(p(v_{j}|v_{i})h^{(l-1)}(v_j)W^{(l)}-F^{(l)}(v_i)q_{j}^{(l-1)})^2}{{n_{i}^{(l-1)}}(q_{j}^{(l-1)})^2}]\nonumber\\
        =&\mathbb{E}_{q_{j}^{(l-1)}}[\frac{(\hat{A}_{ij}h^{(l-1)}(v_j)W^{(l)}-F^{(l)}(v_i)q_{j}^{(l-1)})^2}{n_{i}^{(l-1)}(q_{j}^{(l-1)})^2}]
    \label{(25)}
\end{align}

\subsection{Proof of Theorem 3}
\label{a3}
In order to derive the closed-form solution of minimum variance layer-wise node sampler, we let
$\frac{\partial L(\lambda_{1}, \lambda_{2},\cdots,\lambda_{L},q^{(l-1)}(v_j |{V}^{(l)}))}{\partial q^{(l-1)}(v_j|\hat{V}^{(l)})} = 0$ from Eq. (\ref{(10)}) as

\begin{align}
    &\frac{\partial L(\lambda_{1}, \lambda_{2},\cdots,\lambda_{L},q^{(l-1)}(v_j |{V}^{(l)}))}{\partial q^{(l-1)}(v_j|\hat{V}^{(l)})} \nonumber\\
    =& \frac{\partial \sum_{i=1}^{N}{Var}_{q^{(l-1)}(v_j |{V}^{(l)})} (\hat{F}^{(l)}(v_i))}{\partial q^{(l-1)}(v_j|\hat{V}^{(l)})} - \lambda_l\nonumber\\
    =& \frac{\partial \sum_{i \in V^{(l)}}{Var}_{q^{(l-1)}(v_j |{V}^{(l)})} (\hat{F}^{(l)}(v_i))}{\partial q^{(l-1)}(v_j|\hat{V}^{(l)})} - \lambda_l\nonumber\\
    = & \frac{\partial \sum_{i \in V^{(l)}}\mathbb{E}_{q}[\frac{(\hat{A}_{ij}h^{(l-1)}(v_j)W^{(l)}-F^{(l)}(v_i)q^{(l-1)}(v_j|\hat{V}^{(l)}))^2}{n^{(l-1)}(q^{(l-1)}(v_j|\hat{V}^{(l)}))^2}]}{\partial q^{(l-1)}(v_j|\hat{V}^{(l)})}\nonumber\\
    - &\lambda_l\nonumber\\
    =& \frac{\partial \sum_{i \in V^{(l)}}\sum_{k=1}^{N} [\frac{(\hat{A}_{ik}h^{(l-1)}(v_k)W^{(l)}-F^{(l)}(v_i)q^{(l-1)}(v_k|\hat{V}^{(l)}))^2}{n^{(l-1)}q^{(l-1)}(v_k|\hat{V}^{(l)})}]}{\partial q^{(l-1)}(v_j|\hat{V}^{(l)})}\nonumber \\
    -& \lambda_l\nonumber\\
    =&\sum_{i \in V^{(l)}}\frac{-(\hat{A}_{ij})^2|h^{(l-1)}(v_j)W^{(l)}|^{2}+|F^{(l)}(v_i)|^{2}{q^{(l-1)}(v_j|\hat{V}^{(l)})}^2}{n^{(l-1)}{q^{(l-1)}(v_j|\hat{V}^{(l)})}^2}\nonumber\\
    -& \lambda_l\nonumber\\
    =&0
    \label{(26)}
\end{align}

From Eq. (\ref{(26)}), we can obtain

\begin{equation}
    \begin{split}
    {q^{*}}^{(l-1)}(v_j|\hat{V}^{(l)}) = \frac{\sqrt{\sum_{i \in V^{(l)}}\hat{A}^{2}_{ij}}|h^{(l-1)}(v_j)W^{(l)}|}{\sqrt{\sum_{i \in V^{(l)}}|F^{(l)}(v_i)|^{2} - n^{(l-1)} \lambda_l}}
    \label{(27)}
    \end{split}
\end{equation}

Sum both sides of Eq. (\ref{(27)}) at the same time as

\begin{align}
    &\sum_{j=1}^{N}{q^{*}}^{(l-1)}(v_j|\hat{V}^{(l)}) \nonumber\\
    & \ \ \ \ = \frac{\sum_{j=1}^{N} \sqrt{\sum_{i \in V^{(l)}}\hat{A}^{2}_{ij}}|h^{(l-1)}(v_j)W^{(l)}|}{\sqrt{\sum_{i \in V^{(l)}}(F^{(l)}(v_i))^{2} - n^{(l-1)} \lambda_l}}=1
    \label{(28)}
\end{align}

Then we can find that $\sqrt{\sum_{i \in V^{(l)}}(F^{(l)}(v_i))^{2} - n^{(l-1)} \lambda_l} = \sum_{j=1}^{N} \sqrt{\sum_{i \in V^{(l)}}\hat{A}^{2}_{ij}}|h^{(l-1)}(v_j)W^{(l)}| $. By substituting this in Eq. (\ref{(27)}), we finally attain the variance minimum layer-wise node sampler shown in Eq. (\ref{(12)}). Note that the variance minimum node-wise sampler and the subgraph node sampler shown in Eq. (\ref{(20)}) can be obtained with this derivation process as well.

\subsection{Proof of Theorem 4}
\label{a4}
In subgraph sampling, ${Loss}(s_{i})$  is an unbiased losswhen function of $Loss$ when $\mathbb{E}_{q(v_j)}({Loss}(s_{i}))=Loss$. Similar to the proof of theorem 1, the proof is demonstrated as follows:

\begin{align}
        &\mathbb{E}_{q(v_j)}({Loss}(s_{i}))\nonumber\\
        =&\mathbb{E}_{q(v_j)}(\frac{1}{|V_{t}| \cdot n_{s_{i}}}\sum_{j=1}^{n_{s_{i}}}\frac{1}{q(v_{j})}\mathcal{L}(\hat{y}(v_j), y(v_j)))\nonumber\\
        =&\frac{1}{|V_{t}|}\mathbb{E}_{q(v_j)}(\frac{1}{q(v_{j})}\mathcal{L}(\hat{y}(v_j), y(v_j)))\nonumber\\
        =&\frac{1}{|V_{t}|}\sum_{j=1}^{|V_{t}|}\mathcal{L}(\hat{y}(v_j), y(v_j))\nonumber\\
        =&Loss
    \label{(29)}
\end{align}

\section{Configurations}
\label{b}

Table \ref{t4} summarizes the configurations of our proposed three samplers with highest average "F1-micro" score of 10 trials on validation set. Note that (1) we let the Bernoulli sampling number equivalent with the batch size in layer-wise sampling and (2) batch size in subgraph sampling denotes the number of Bernoulli sampling, which implies that the scale of the actual computation graph is slightly smaller than this batch size due to node resampling.

\end{document}